\documentclass[twoside,11pt]{jmlr} 

\usepackage{times}

\author{\Name{Kamil Nar}
\hfill \includegraphics[trim={0, 2pt, 0, 0}, clip, scale=0.1825]{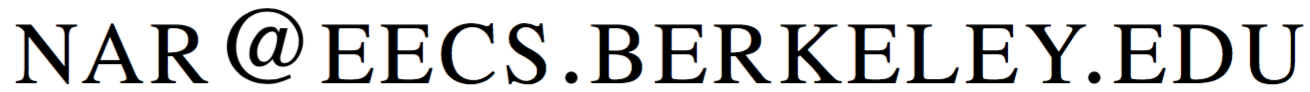}\\
 \Name{S. Shankar {Sastry}} \hfill \includegraphics[trim={0, 2pt, 0, 0}, clip, scale=0.1825]{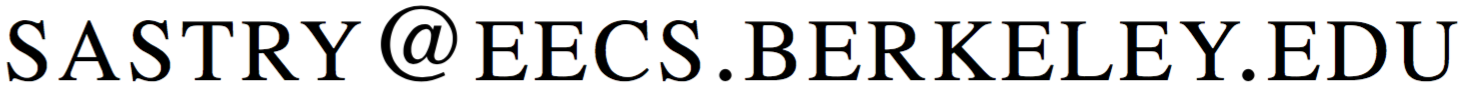}\\
 \addr Electrical Engineering and Computer Sciences, University of California, Berkeley}
 
\usepackage{graphicx}				
\usepackage{float}
\usepackage{amssymb}
\usepackage{amsmath}
\usepackage{dsfont}
\usepackage{centernot}
\usepackage{euscript}
\usepackage{diagbox}

\usepackage{multirow}
\usepackage{booktabs}
\usepackage{pgfplotstable}
\usepackage{pgfplots}
\usepackage{graphicx}
\usepackage{tikz}

\usepackage{algorithm}
\usepackage[noend]{algpseudocode}

\DeclareMathOperator*{\argmax}{argmax}
\DeclareMathOperator*{\argmin}{argmin}
\DeclareMathOperator*{\subjectto}{subject\ to}

\usepackage{lipsum}

\newtheorem{theo}{Theorem}

\newtheorem*{defn}{Definition}

\title{Persistency of Excitation for
Robustness of Neural Networks
}

\begin{document}

\maketitle

\begin{abstract}
When an online learning algorithm is used to estimate the unknown parameters of a model, the signals interacting with the parameter estimates should not decay too quickly for the optimal values to be discovered correctly.
This requirement is referred to as persistency of excitation, and~it arises in various contexts, such as optimization with stochastic gradient methods, exploration for multi-armed bandits, and adaptive control of dynamical systems.
While training a neural network, the iterative optimization algorithm involved also creates an online learning problem, and~consequently, correct estimation of the optimal parameters requires persistent excitation of the network weights.
In this work, we analyze the dynamics of the gradient descent algorithm while training a two-layer neural network with two different loss functions, the squared-error loss and the cross-entropy loss; and we obtain  conditions to guarantee persistent excitation of the network weights.
We then show that these conditions are difficult to satisfy when a multi-layer network is trained for a classification task, for the signals in the intermediate layers of the network become low-dimensional during training and fail to remain persistently exciting.
To provide a remedy, we delve into the classical regularization terms used for linear models, reinterpret them as a means to ensure persistent excitation of the model parameters,
and propose an algorithm for neural networks by building an analogy. 
The results in this work shed some light on why adversarial examples have become a challenging problem for neural networks, why merely augmenting training data sets will not be an effective approach to address them, and why there may not exist a data-independent regularization term for neural networks, which involve only the model parameters but not the training data.
\\ \ 
\end{abstract}


\vfill
\noindent\rule{0.5\textwidth}{0.4pt}

\noindent{\footnotesize The implementation of all experiments is available at\ \href{https://github.com/nar-k/persistent-excitation}{ https://github.com/nar-k/persistent-excitation}.}

\newpage 
\tableofcontents

\newpage

\section{Introduction}

State-of-the-art neural networks provide high accuracy for plenty of machine learning tasks, but their performance has proven vulnerable to small perturbations in their inputs \citep{Szegedy}. 
This lack of robustness prohibits the use of neural networks in safety-critical applications such as computer security, control of cyber-physical systems, self-driving cars, and medical prediction and planning \citep{kurakin2016adversarial,strickland2019ibm}.

Given the combination of large number of training parameters, nonlinearity of the model, nonconvexity of the optimization problem used for training, involvement of an iterative algorithm for optimization, and high-dimensionality of the commonly used data sets, it is challenging to pinpoint what particularly causes the lack of robustness in neural networks. It has been speculated that the high nonlinearity of neural networks gives rise to this vulnerability \citep{Szegedy}, but the fact that support vector machines with nonlinear feature mappings remain much more robust against similar perturbations readily invalidates this claim \citep{Goodfellow2015Adversarial}.

It has been observed that the lack of robustness is not a consequence of the depth of the network; even networks with few layers are shown to produce drastically different outputs for almost identical inputs. Comparison of this observation with the robustness of support vector machines has led to the belief that neural networks might in fact not be introducing sufficient level of nonlinearity to execute the learning tasks involved --- since shallow networks function as an affine mapping in most of their domains, whereas most feature mappings used in support vector machines are unarguably nonlinear \citep{Goodfellow2015Adversarial}. Although this belief still prevails, the level of nonlinearity is not the only aspect that neural networks and support vector machines differ in. There are many~other factors these two structures do not share in common, 
and these factors might also account for their dissimilarity in robustness. 

One of the important yet overlooked differences between neural networks and support \hbox{vector} machines is the first-line loss functions used in their training. For classification tasks, neural networks are almost always trained with the cross-entropy loss, whereas training support vector machines involves the hinge loss function \citep{hastie_09_elements-of.statistical-learning}.
This raises the question whether the choice of training loss function, and in particular the cross-entropy loss function, is one of the factors leading to the lack of robustness in neural networks.
To provide a preliminary answer to this question, we train a two-layer neural network for a binary classification task with two different loss functions: the cross-entropy loss and the squared-error loss. Figure 1 demonstrates the decision boundaries of the network trained with the two loss functions for different initializations.
Even though the network architecture, the training data, and the optimization algorithm are kept identical, Figure~1 shows that the network trained with the cross-entropy loss consistently has a much smaller margin than the network trained with the squared-error loss, and hence, it is less robust. 

\begin{figure}[h!]
    \centering
    \includegraphics[width=0.361\textwidth]{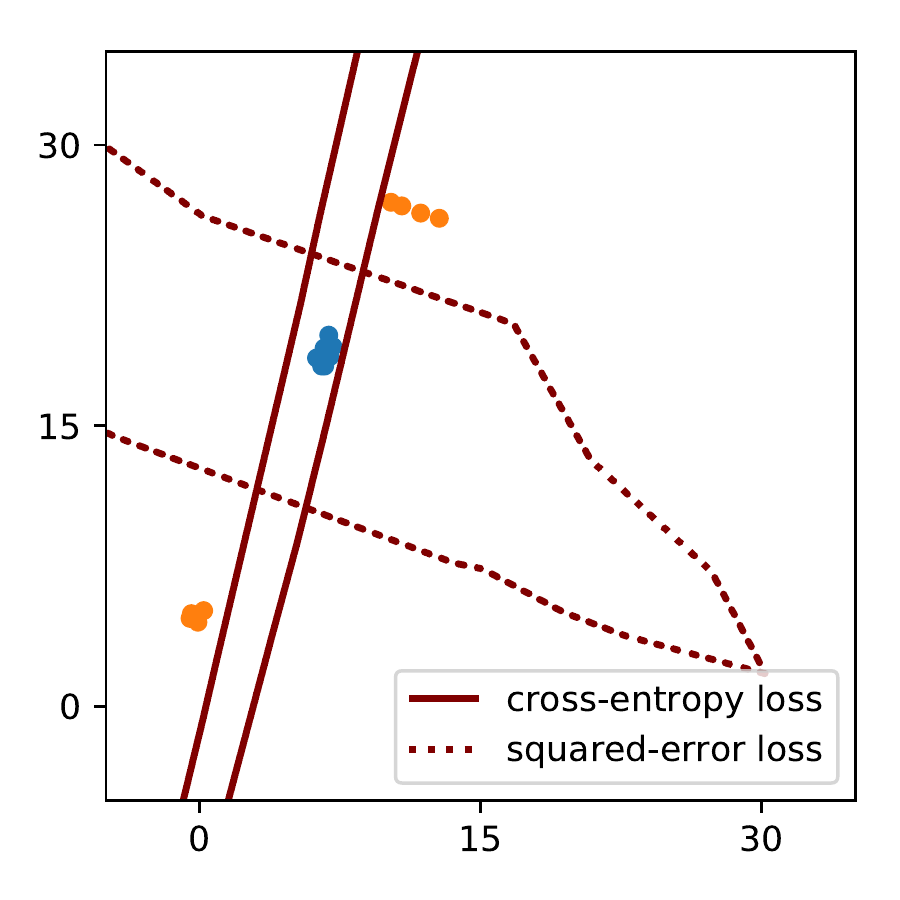}
    \includegraphics[width=0.361\textwidth]{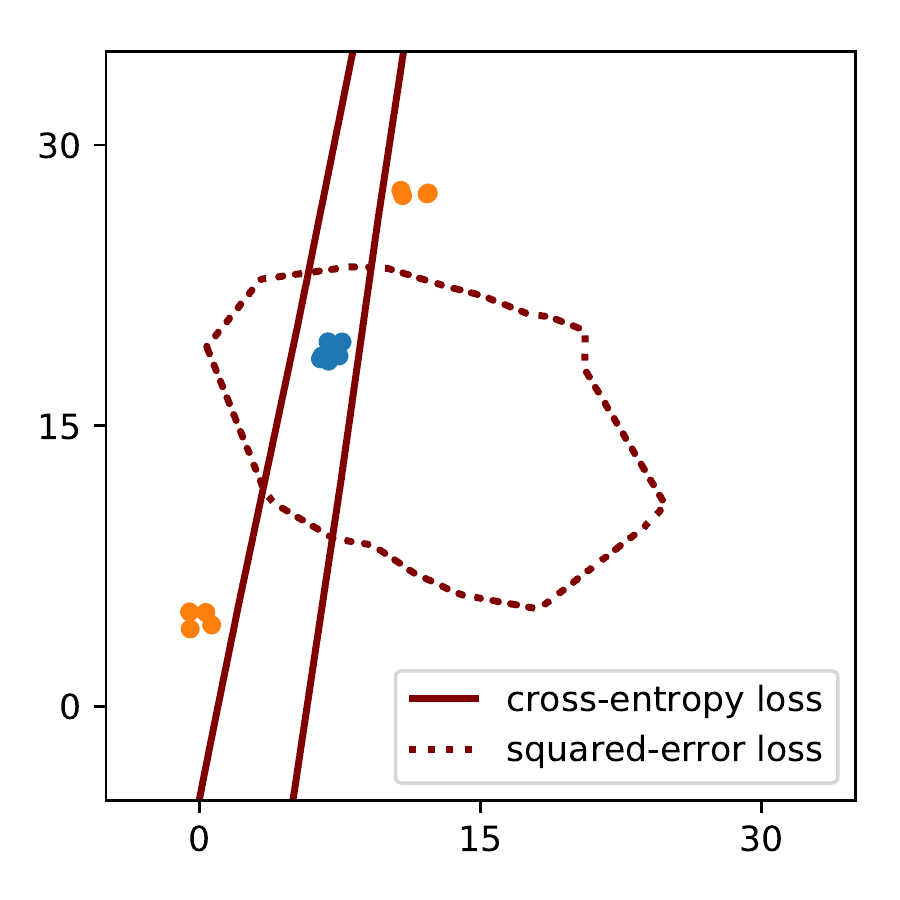}\\
    \includegraphics[width=0.361\textwidth]{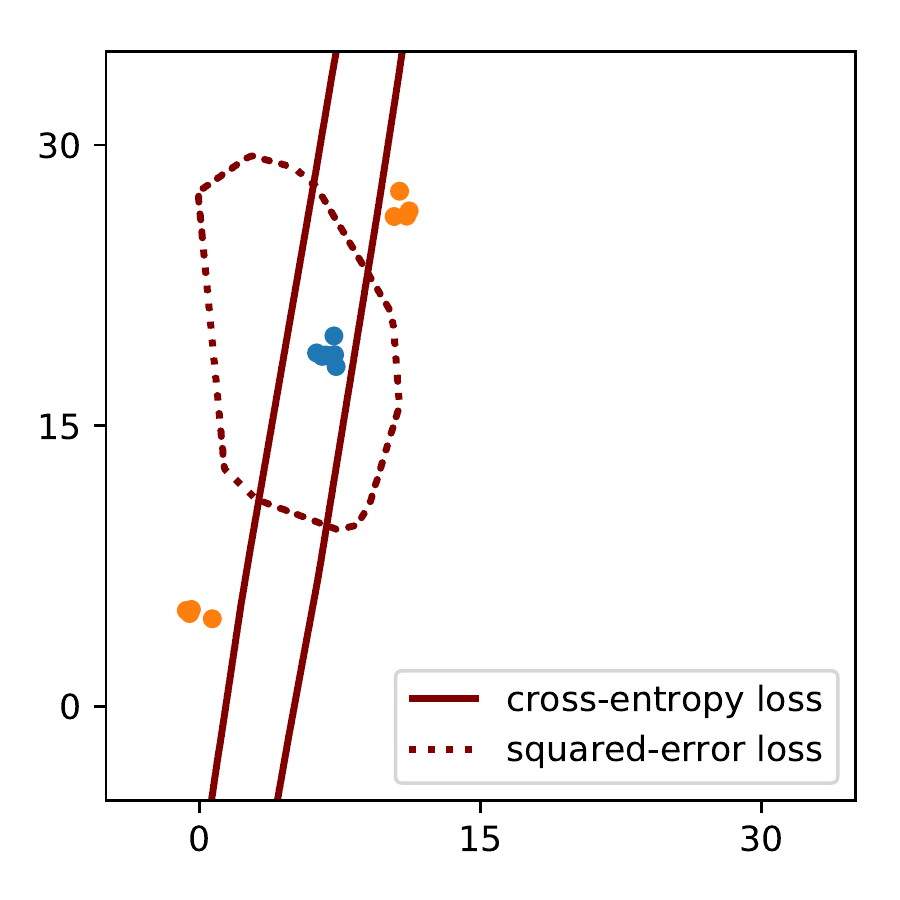}
    \includegraphics[width=0.361\textwidth]{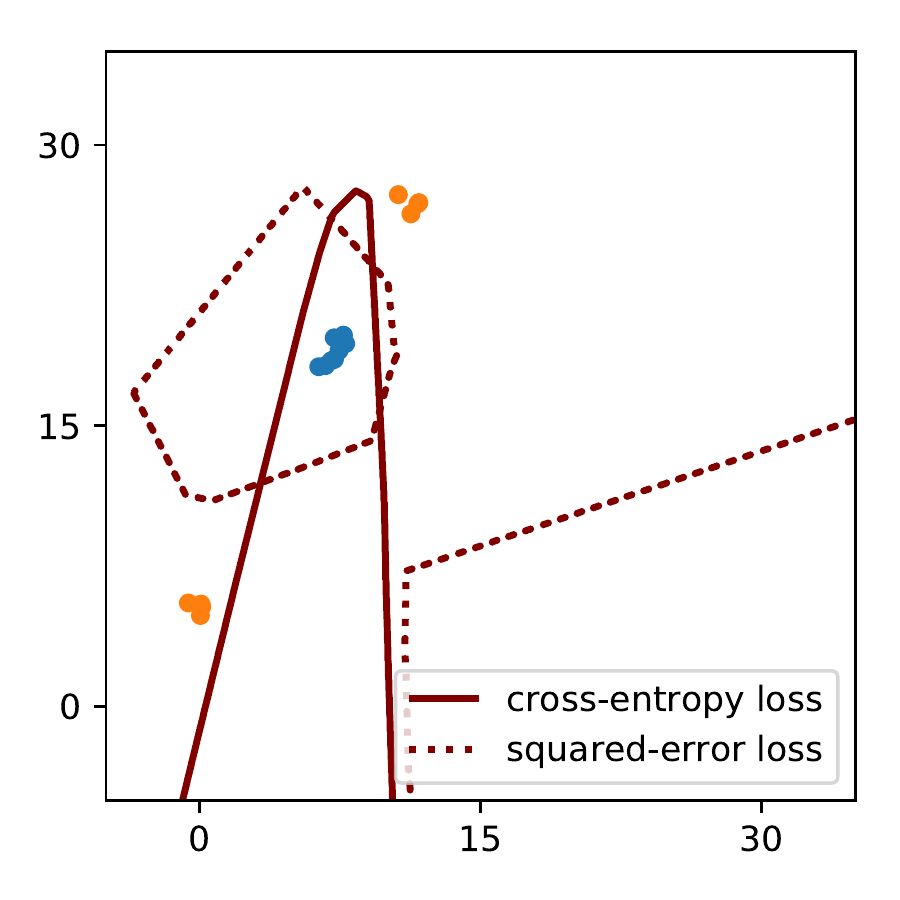}\\
    \includegraphics[width=0.361\textwidth]{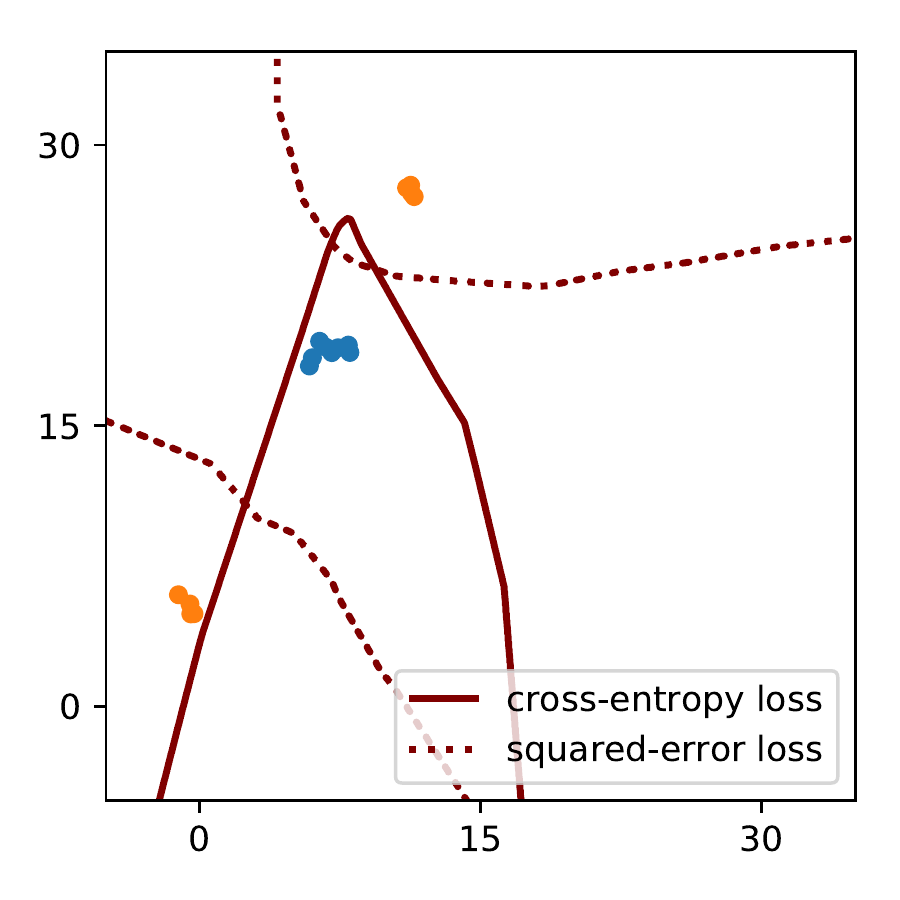}
    \includegraphics[width=0.361\textwidth]{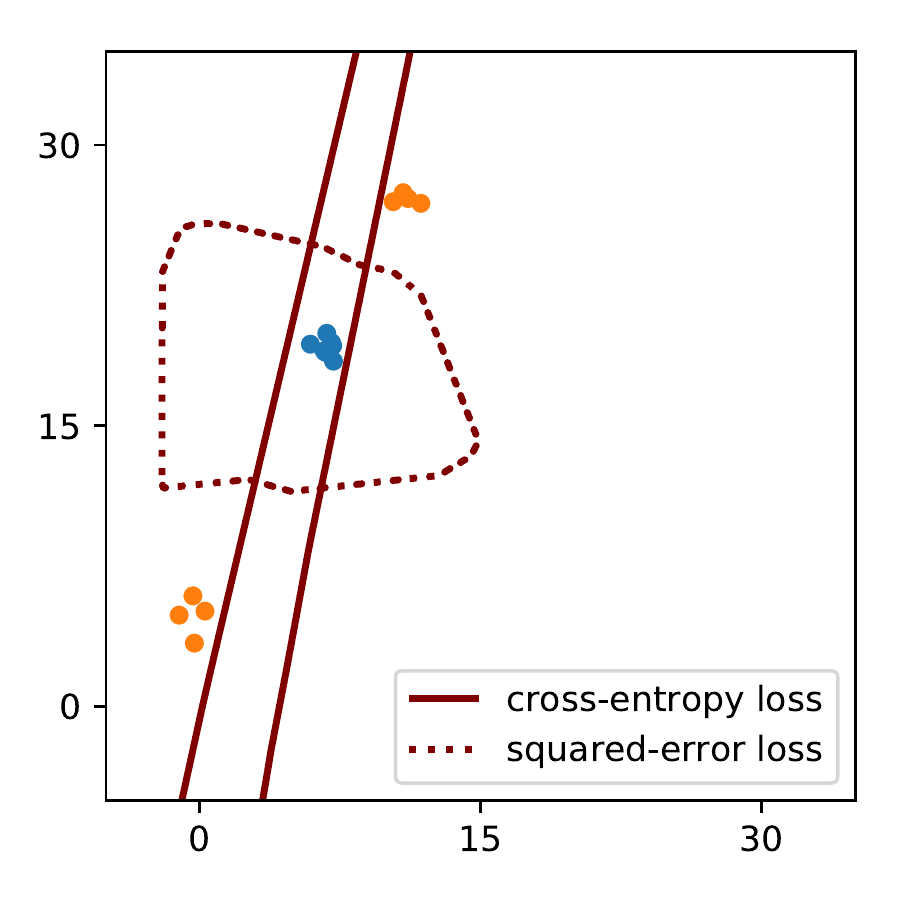}
    \caption{The decision boundaries of a two-layer network trained with the cross-entropy loss and the squared-error loss.  The task is binary classification in $\mathbb R^2$; the orange clusters form one class and~the blue cluster represents the other class. Each plot shows the decision boundaries for a different initialization, corresponding to the random seeds 1, 21, 41, 61, 81, 101. The network architecture, the optimization algorithm, and the training data are identical within each plot; only the training loss functions are different. The networks trained with the cross-entropy loss consistently have a substantially poorer margin.}
\end{figure}


Figure 1 confirms that at least some ingredients of neural network training procedure have an influence on the robustness of networks.
Naturally, the next question to ask is whether the identical training procedure could lead to a similar vulnerability in models that are different than neural networks.
Figure 2 illustrates two examples of a linear classifier trained with the cross-entropy loss function and the gradient descent algorithm.
It appears that the decision boundary of a linear classifier could also have an extremely poor margin when it is trained in exactly the same way as neural networks are trained.

\begin{figure}
    \centering
    \begin{minipage}{0.5\textwidth}
    \includegraphics[trim={0, 0.1in, 0, 0in}, clip, scale=0.7]{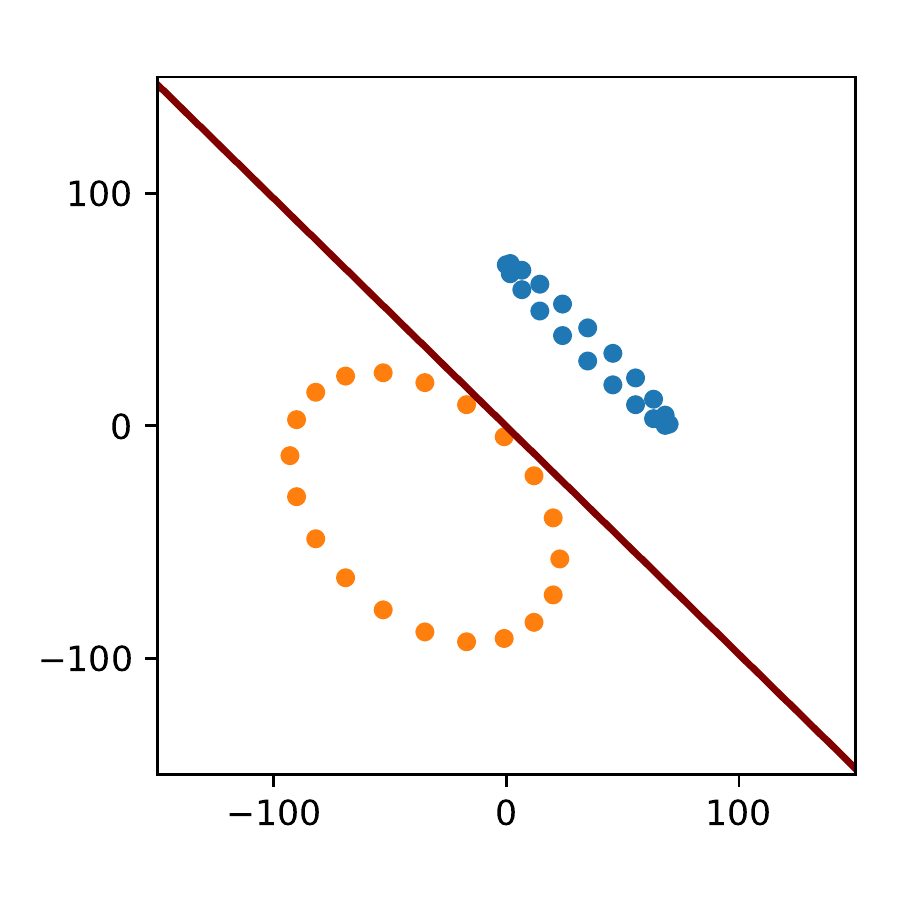}
    \end{minipage}%
    \begin{minipage}{0.5\textwidth}
    \includegraphics[trim={0, 0.1in, 0.1in, 0in}, clip, scale=0.7]{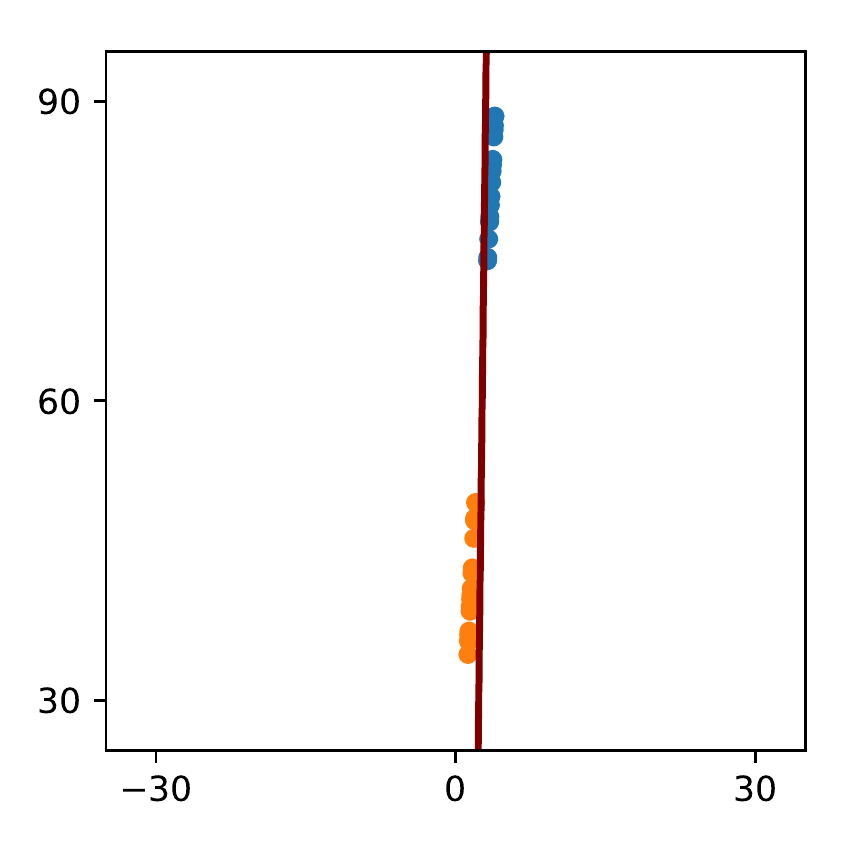}
    \end{minipage}
    \caption{Two different data sets on which a linear classifier is trained by minimizing the cross-entropy loss using the gradient descent algorithm. The solid lines represent the decision boundary of the classifier, which lie very close to the training data. Standard ingredients of neural network training seem to cause extremely poor margin and lack of robustness for linear classifiers as well.}
    \label{fig:poor-margins}
\end{figure}

To understand the observations in Figure 1 and Figure 2, we will first look into a concept called persistency of excitation,
and then explain how it relates to training neural networks.

\subsection{Persistency of Excitation}

Consider the training of a linear model with the squared-error loss:
\[ \min_{\theta} \sum\nolimits_{i \in \mathcal I} {\| x_i^\top  \theta - y_i \|}_2^2, \]
where $\{x_i\}_{i \in \mathcal I}$ is the set of training data in $\mathbb R^n$ and $\{y_i\}_{i \in \mathcal I}$ is the set of target values in $\mathbb R$.
This problem is convex, and if the set $\{x_i\}_{i \in \mathcal I}$ spans whole $\mathbb R^n$, the optimal value for $\theta$ will be unique. We can use the gradient descent algorithm to find this optimal value:
\begin{equation}
\label{upd:theta} \theta \gets \theta - \delta \sum\nolimits_{i \in \mathcal I} x_i (x_i^\top \theta - y_i), \end{equation}
where $\delta$ is some fixed step size.
As long as $\delta$ is sufficiently small, the parameter estimate is guaranteed to converge to the optimal value of $\theta$.

Now, assume for some reason the measurements are attenuated at each iteration of the gradient descent algorithm. That is, the update rule (\ref{upd:theta}) is followed at each iteration by
\[
(x_i,y_i) \gets (\alpha x_i, \alpha y_i) \quad \forall i\in \mathcal I, \]
for some $\alpha \in (0,1)$. In this case, the gradient descent algorithm will lose its ability to find the true value of $\theta$. For every initialization, the algorithm will still converge, the error term will still become zero, but the parameters will not necessarily converge to the optimal $\theta$. 

This might seem like a contrived example, but it is a fundamental problem that arises in various settings, such as identification and adaptive control of dynamical systems, optimization with stochastic gradient methods,
and exploration in multi-armed bandits.

\begin{itemize}
    \item \textbf{System identification and adaptive control:} Consider a discrete-time linear dynamical system:
    \begin{equation}
    \label{dynamic:sys}
    x_{t+1} \gets \theta x_t + u_t \quad \forall t \in \mathbb N,\end{equation}
    where $x_t \in \mathbb R$ and $u_t \in \mathbb R$ are the state and the input of the system at time $t$, and $\theta \in (0,1)$ is the unknown parameter of the system. Assume an estimator system of the form
    \[ \hat x_{t+1} \gets \hat \theta \hat x_t + u_t \quad \forall t \in \mathbb N \]
    is used to identify the value of $\theta$ by decreasing the distance between $x_t$ and $\hat x_t$ over time with the gradient descent algorithm. Then both $x_t$ and $\hat x_t$ can decay to zero while $\hat \theta$ converges to some value different than $\theta$ \citep{kumar2015stochastic}. 
    
    \item \textbf{Stochastic gradient methods:} Consider a convex function $f(\theta)$, which can be decomposed as $\sum_{i \in \mathcal I} f_i(\theta)$, where each $f_i$ is also a convex function. Assume we use the following stochastic gradient method with a fixed step size $\delta$:
    \begin{enumerate}
        \item Randomly choose $i \in \mathcal I$,
        \item $\theta_{t+1} \gets \theta_t - \delta {\partial \over \partial \theta} f_i(\theta)$,
        \item Scale down $f(\theta)$; in other words, set $f_i \gets \alpha_t f_i$ for all $i \in \mathcal I$ for some $\alpha_t \in (0,1)$,
        \item Return to 1.
    \end{enumerate}
    This algorithm does not necessarily converge to the optimal value for $\theta$ if $\alpha_t$ attenuates $f(\theta)$ too quickly \citep{bottou2018optimization}.
    \item \textbf{Multi-armed bandits:} Consider a two-armed bandit problem where the arms produce the independent and identically distributed random processes $\{X_t\}_{t \in \mathbb N}$ and $\{\tilde X_t\}_{t \in \mathbb N}$ with distinct means; that is, $\mathbb E X_t \neq \mathbb E \tilde X_t$. Assume we try to solve
    \[ \max_{ \{d_t\}_{t \in \mathbb N} \in \{0,1\}^{\mathbb N}} \mathbb E \left( \lim_{T \to \infty}  {1 \over T} \sum_{t=1}^T \left(d_t X_t + (1-d_t) \tilde X_t\right) \right) \]
    by using a causal feedback mechanism $f: \mathbb N^2 \times \mathbb R^2 \mapsto \{0,1\}$ that prescribes a decision at time $t \in \mathbb N$ as:
    \[ d_t = f\left( t, \sum\nolimits_{t'=1}^{t-1} d_{t'},  {\sum\nolimits_{t' =1}^{t-1} d_{t'}X_{t'}},
    {\sum\nolimits_{t' =1}^{t-1} (1-d_{t'})\tilde X_{t'}}
    \right).
    \]
    If the feedback policy $f$ decreases the frequency of pulling the arm with the lower average of observed rewards too quickly, then the algorithm can fail to discover the arm with the higher mean \citep{bubeck2012regret}.
\end{itemize}

In these three examples, there is an input, a signal, or an action that excites the unknown parameters; that is, something that forces the parameters to release some information about themselves. 
We observe that when this excitation decays too quickly, the online learning algorithm cannot receive necessary amount of information about the parameters and fail to estimate them correctly.
This leads us to a concept~called \emph{persistency of excitation}. For online algorithms to learn the unknown parameters of a model correctly, the signals interacting with the parameter estimates need to remain persistently exciting during the estimation procedure \citep{kumar2015stochastic, sastry2011adaptive}. If this persistency of excitation condition is not satisfied, the error terms inside the learning algorithm can become zero even if the parameters converge to a wrong value or do not converge at all.


\subsection{Connection Between Persistency of Excitation and Neural Networks}

During training of a feedforward neural network, the iterative optimization algorithm involved creates the dynamics of an online learning problem as shown in Figure 3. The function $f_{\theta}$ represents the true mapping with the ideal parameters, whereas $f_{\hat \theta}$ denotes the estimate obtained with the gradient descent algorithm.
Assume that the neural network has  $L$ layers, and it is described as
\[ f_{\theta}(x) = f_{\theta_L} \circ f_{\theta_{L-1}} \circ \cdots  \circ f_{\theta_1} (x), \]
where $f_{\theta_k}$ represents the operation the $k$-th layer of the network performs and $\theta_k$ stands for the parameters of that particular layer for each $k \in [L]$. Given a set of training data $\{x_i\}_{i \in \mathcal I}$, the parameters of the $k$-th layer, $\theta_k$, are excited by the signals of the preceding layer throughout the training, which are given as $\{ f_{\theta_{k-1}} \circ \cdots \circ f_{\theta_1}(x_i) \}_{i \in \mathcal I}$. Note that no matter how large the training data set is, there is no strong reason to think that these signals in the intermediate layers will remain persistently exciting during training, and we will show in this work that they indeed do not remain persistently exciting with standard training procedures.
The consequence is that even if the training error converges to zero, the estimated parameters may not be the same as the ideal parameters of the true mapping, and they may not
predict the output of the network accurately for unseen data.

\begin{figure}[ht]
    \centering
    \includegraphics[trim={3in, 3.3in, 3in, 3.3in}, clip, scale=0.5]{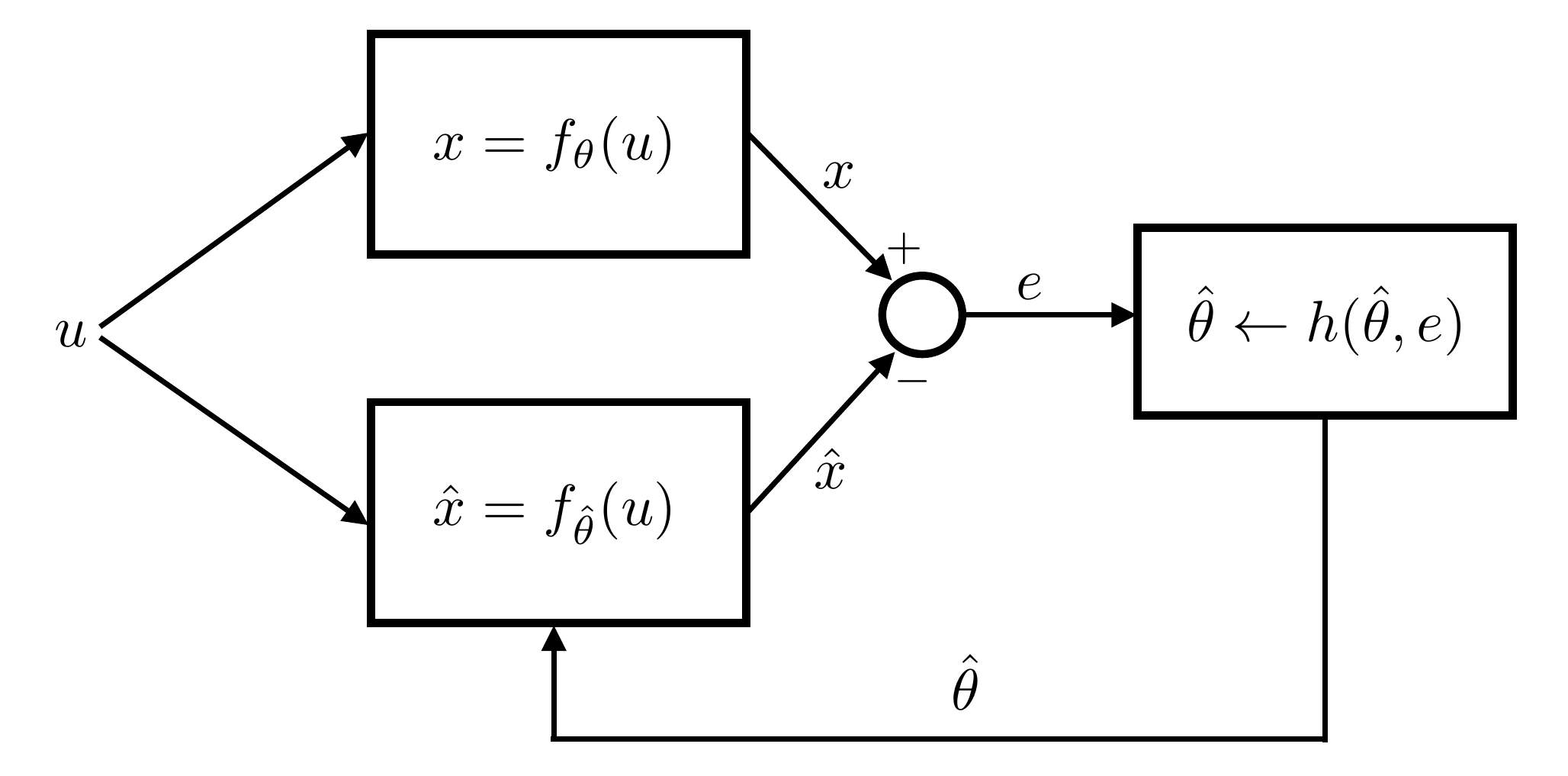}
    \caption{Closed-loop representation of the dynamics of neural network training. Function $f_\theta$ represents the true mapping that we want to estimate, while $f_{\hat \theta}$ represents the mapping estimated via the gradient descent algorithm. Signal $e$ denotes the training error of the model estimated, and function $h$ gives the update rule for the parameter estimates.}
    \label{fig:training-loop}
\end{figure}

We will see in the following sections that there exist some conditions on the training data which ensure the persistent excitation of the network parameters. Given these conditions, convergence of training error to zero will imply the convergence of the estimate $\hat \theta$ to an optimal parameter $\theta$ that will induce the network to produce similar outputs for similar inputs.

The concept of persistency of excitation already provides an interpretation of Figure 1 and Figure~2. Different training loss functions give rise to different dynamics for the gradient descent algorithm; hence, they involve different conditions on the training data for the persistent excitation of the parameters --- some of which are easier to satisfy than the others. Moreover, for some loss functions,  even the linear models are not exempted from the restrictiveness of these conditions.

\subsection{Our Contributions and Outline}
The next section starts with the analysis of the gradient descent algorithm on the squared-error loss function. It provides a richness condition on the training data set such that the convergence of the gradient descent algorithm guarantees a small Lipschitz constant for the estimated model.

Section 3 studies the dynamics of the gradient descent algorithm on the cross-entropy loss function. For linear and nonlinear classifiers, conditions for persistency of excitation are derived and shown to be different than the conditions required for the squared-error loss. This difference will clarify the contrast between the decision boundaries of the two loss functions in Figure 1.

It has been consistently observed in recent works \citep{mahoney2018} that neural~networks generate low-dimensional signals in their hidden layers.
Section 4 provides a simple theoretical analysis confirming this phenomenon for deep linear networks.
This fact suggests the conditions given in Section 2 and Section 3 for persistency of excitation are in fact difficult to satisfy on multiple-layer neural networks trained with the gradient descent algorithm.

To address the problem raised in Section 4, we will look closely into the classical approach that is used to strengthen the correspondence between the training and test performances of a model: adding regularization.
The traditional viewpoint is that adding a regularization term to the training loss function assigns a prior distribution to the parameters of the model and restricts the class of functions that can be represented by the model \citep{hastie_09_elements-of.statistical-learning}.
Section 5 introduces a dual viewpoint: the regularization term creates a richer training data set and eases meeting the condition for the persistency of excitation. By building an analogy with this viewpoint, a new training scheme is introduced to ensure persistent excitation of all parameters of a neural network during its training.

Section 6 contains the experimental results of the algorithm proposed for persistent excitation of neural networks. Section 7 discusses the results and provides a comparison with the related works, and finally Section 8 concludes the paper. Appendices contain the proofs of the theorems and the implementation details for the experiments.

\section{Gradient Descent on Squared-Error Loss}

Given a regression problem with the squared-error loss, it is natural to expect that the Lipschitz constant of the estimate will depend on the Lipschitz constant of the actual function generating the target values for the input points. Therefore, one might anticipate determining the Lipschitz constant of the estimate will require information about the relationship between the input points and their target values.
This indeed is the case for linear estimators, but not for neural networks.
The following subsections will show that when a neural network is trained with the gradient descent algorithm, the Lipschitz constant of the estimate will have a bound depending only on the set of training points but \emph{not} on their target values. In other words, the Lipschitz constant of the estimate will be restricted no matter how sharply the actual function changes in its domain.

\subsection{Lack of Input-Based Lipschitz Bounds for Linear Estimators}

Consider again the problem of finding a linear estimator for a function by minimizing its squared-error loss over a set of points $\{x_i\}_{i \in \mathcal I}$.
Let $f : \mathbb R^n \to \mathbb R^m$ be the function that is being estimated, and let  $\ell: \mathbb R^{m \times n} \to \mathbb R$ denote the loss function that is being minimized:
\[ \ell(W) = {1\over 2}\sum\nolimits_{i \in \mathcal I} \left\| W x_i - f(x_i)
\right\|_2^2.\]
If we use the gradient descent algorithm with a fixed step size $\delta$ to minimize $\ell$, the update rule for $W$ becomes
\begin{align*} W 
& \gets W \left( I - \delta \sum\nolimits_{i \in \mathcal I} x_i x_i^\top \right) + \delta \sum\nolimits_{i \in \mathcal I} f(x_i)x_i^\top, 
\end{align*}
which is a discrete-time linear time invariant system. Therefore, a necessary and sufficient condition for the~convergence of the algorithm is that \citep{callier2012linear}
\begin{equation}
\label{eqn:lip1}
 \lambda_\text{max} \left( \sum\nolimits_{i \in \mathcal I} x_i x_i^\top \right) < {2 \over \delta}. 
 \end{equation}
The left hand side of (\ref{eqn:lip1}) is the largest eigenvalue of the covariance matrix of the training data, which is nothing but the Lipschitz constant of the gradient of the loss function $\ell$.

This result is well-known \citep{bertsekas1999nonlinear}: when a differentiable function $\ell : \mathbb R^n \to \mathbb R$ is minimized by using the gradient descent algorithm with a fixed step size $\delta$, the algorithm can converge to a local minimum $w_0 \in \mathbb R^n$ only if the Lipschitz constant of the gradient of $\ell$, $L(\nabla \ell)$, at that point satisfies
\[ L \left( \nabla \ell(w_0) \right) < {2 \over \delta}. \]
This condition for convergence establishes a relationship between the smoothness parameter of the loss function and the step size of the algorithm.

The inequality (\ref{eqn:lip1}), however, does \emph{not} have any implication on the Lipschitz constant of \emph{the estimate for $f$} --- which is the largest singular value of $\hat W$. Indeed, the spectral radius of $\hat W$ could be arbitrarily large; it does not require the step size to be small for convergence. In contrast, Theorem~1 in the following subsection shows that when a neural network is being trained with the gradient descent algorithm, the step size does give information about the Lipschitz constant of the function estimate.
This forms  the crucial difference between the result of Theorem 1 and the well-known relationship between the step size and \emph{some} Lipschitz constant.

\subsection{Input-Based Lipschitz Bounds for Two-Layer Neural Networks}

Now consider a two-layer neural network with ReLU activation in its hidden layer:
\[ x \mapsto W(Vx + b)_+ \]
where $W\in \mathbb R^{m \times r}$, $V\in \mathbb R^{r\times n}$ and $b \in \mathbb R^{r}$. The following theorem shows that if this network~is trained to estimate a function $f : \mathbb R^n \to \mathbb R^m$ by minimizing the squared error loss via gradient descent algorithm, then the convergence of the algorithm implies Lipschitzness of the function estimate $\hat f(x) = \hat W(\hat V x + \hat b)_+$ as long as the training data set is rich enough.


\begin{theo}
\label{theo:theo-1}
Given a function $f: \mathbb R^n \to \mathbb R^m$ and a set of points $\{x_i\}_{i \in \mathcal I}$ in $\mathbb R^n$, assume that a two-layer neural network with parameters $W \in \mathbb R^{m \times r}$, $V \in \mathbb R^{r\times n}$, $b \in \mathbb R^r$ and ReLU activations is trained by minimizing the squared error loss 
\begin{equation}
\label{eqn:squared-loss} \min_{W,V,b} \ {1\over 2}  \sum\nolimits_{i \in \mathcal I} {\left\| W(Vx_i + b)_+ - f(x_i) \right\|}_2^2 \end{equation}
via the gradient descent algorithm. 
Assume that the algorithm has converged from a random initialization\footnote{For this theorem, and all other theorems in this paper, the distribution used for random initialization is assumed to assign zero probability to every set of Lebesgue measure zero.} to a stationary point $(\hat W, \hat V, \hat b)$ with a fixed step size $\delta$. Let $\hat V_k$ and $\hat b_k$ denote the $k$-{th} rows of $\hat V$ and $\hat b$ at equilibrium, and define the set of training points that activate the $k$-{th} node in the hidden layer as
\[\mathcal I_k = \big\{ i \in \mathcal I : \hat V_k x_i + \hat b_k > 0 \big\}.\]
Then the Lipschitz constant of the function estimate $\hat f(x) = \hat W(\hat Vx + \hat b)_+$ is almost-surely upper bounded by
\begin{equation}
\label{bound:main}
\mathcal L(\delta, \overline \mu, \underline \lambda, \hat b, n^\text{\emph{max}}_\text{\emph{active}} ) :=
 n^\text{\emph{max}}_\text{\emph{active}} \left( \sqrt{2 \over \delta \underline \lambda} \right) \left( {2 \overline \mu \|\hat b\|_\infty \over \underline \lambda} + \sqrt{{1 \over \underline \lambda}\left| {2 \over \delta} - \|\hat b\|_\infty^2\right| }
\right)
\end{equation}
where 
\begin{itemize}
\item $n^\text{\emph{max}}_\text{\emph{active}}$ is the maximum number of hidden-layer nodes that a point in $\mathbb R^n$ can activate:
\[ n^\text{\emph{max}}_\text{\emph{active}} = \max_{x \in \mathbb R^n} \ \sum\nolimits_{k=1}^r \mathbf{1}_{\left\{ \hat V_k x + \hat b_k > 0 \right\}}, \]
\item $\underline \lambda$ is a lower bound for the minimum-eigenvalue of the covariance matrix of the training points that activate the same hidden-layer node:
\[ \underline \lambda = \min_{k\in [r]} \ \lambda_{\text{\emph{min}}} \Big( \sum\nolimits_{i \in \mathcal I_k} x_i x_i^\top
\Big),
\]
\item $\overline \mu$ is an upper bound for the $\ell_2$ norm of the summation of the training points that activate the same hidden-layer node:
\[ \overline \mu = \max_{k \in [r]} \  \Big\| \sum\nolimits_{i \in \mathcal I_k} x_i
\Big\|_2. \]
\end{itemize}
\end{theo}

\proof{ See Appendix \ref{app:theo-1}. \hfill \BlackBox}

Theorem 1 underlines a fact: convergence of the training algorithm guarantees a Lipschitz bound for the network, from its input to its output, provided that every node in the hidden layer is excited from every dimension, i.e., the set of points that activate each hidden layer node spans the whole input space. This provides a sufficient condition for the Lipschitzness of the network in terms of the richness of the training data.

Theorem 1 also reveals the following:
\begin{enumerate}
    \item Convergence with a larger step size implies a smaller Lipschitz constant for the mapping represented by the network, from its input to its output. This extends the similar result shown for deep linear networks in \citep{nipsStepSize} to two-layer nonlinear networks.
    \item If the set of points activating any one of the hidden layer nodes has a large bias, then $\overline \mu$  will be large, and so will the upper bound for the Lipschitz constant.
    \item As the number of hidden-layer nodes increases, so does the bound for the Lipschitz constant. This might seem to contradict with the premise of \citep{bartlett1998sample}, which establishes that the weights of the network parameters are more important than the size of the network for
    its performance on training and test data sets to be similar. However, \cite{bartlett1998sample} studies a network with its parameters specified a priori --- independently of whether those weights could be obtained by an iterative algorithm or not. Theorem 1, on the other hand, shows that if the size of the network is large, the gradient descent algorithm can potentially drive the network weights in such a way that the Lipschitz constant of the network becomes large.
\end{enumerate}

If a network is trained for a classification task, the small Lipschitz constant of the network, from its input to its output,  will suggest a large margin between the training data and the decision boundary of the classifier. This is formalized in the following corollary.

\begin{corollary} Given two sets of points $\{x_{i}\}_{i \in \mathcal I}$ and $\{x_{j}\}_{j \in \mathcal J}$ in $\mathbb R^n$, assume that a two-layer neural network is trained to classify them by minimizing the squared-error loss:
\[ \min_{W, V, b} \ {1 \over 2} \sum\nolimits_{i \in \mathcal I} {\| W(Vx_{i} + b)_+ - 0 \|}_2^2
+ {1 \over 2} \sum\nolimits_{j \in \mathcal J} {\| W(Vx_{j} + b)_+ - 1 \|}_2^2
 \]
via the gradient descent algorihtm. If the algorithm converges from a random initialization to a solution $\hat f(x) = \hat W(\hat Vx + \hat b)_+$ with a step size $\delta$, and if 
\[ \Delta := \min\nolimits_{j \in \mathcal J} \hat f(x_{j}) - \max\nolimits_{i \in \mathcal I} \hat f(x_{i}) > 0, \]
then the classifier
\[ x \mapsto \text{\emph{sign}}\left( \hat f(x) - \max\nolimits_{i \in \mathcal I} \hat f(x_i) - {\Delta/2}
\right)
\]
has a margin (in the input space) greater than or equal to
\[ {\Delta \over 2 \mathcal L(\delta, \overline \mu, \underline \lambda, b, n^\text{max}_\text{active} )} \]
almost surely, where $\mathcal L$ is as defined in (\ref{bound:main}). \hfill \BlackBox

\end{corollary}
 The same relationship between a small Lipschitz constant and a large margin will hold for the similar results in the following sections.

\section{Gradient Descent on Cross-Entropy Loss}
\label{section:cross-entropy}

Logistic regression and its counterpart for multiple classes, cross-entropy minimization with the soft-max function, are widely used for classification tasks.
In this section, we analyze the dynamics of the gradient descent algorithm on these loss functions and show how the concept of persistent excitation arises if a guarantee for a large margin is needed for the classifier.

Note that state-of-the-art neural networks are observed to achieve zero training error for classification tasks \citep{RechtUnderstanding}, which implies that the features of the training points eventually become linearly separable in some layer of the networks during the training procedure.
For this reason, we focus our analysis on the case where the features of the training data points are linearly separable.

\subsection{Margin Analysis for Linear Classifier}

When a linear classifier is trained with logistic regression for a binary classification task, the optimization problem involved is known to be convex. Nevertheless, this optimization problem cannot attain its minimum if the training points are linearly separable, that is, if there is a hyperplane separating the points of one class from the others \citep{hastie_09_elements-of.statistical-learning}. The objective function could be made arbitrarily close to zero by driving the parameters to infinity in norm. Then the direction of the growth of the parameters becomes crucial, since this direction determines the normal vector of the separating hyperplane, and consequently, the margin of the classifier.

In recent work of \cite{Soudry-March-2018}, it has been shown that when the training data set is linearly separable, the separating hyperplane obtained with cross-entropy minimization via the gradient descent algorithm is described by a support vector machine problem in an augmented space ---  with the caveat that the hyperplane is constrained to pass through the origin in this augmented space. For this reason, the margin obtained with the cross-entropy minimization is suboptimal in the original space the training points lie in. The following theorem demonstrates how this margin compares to the true optimal margin.

\begin{theo}
\label{theo:theo-2}
Assume that the points $\{x_{i}\}_{i\in \mathcal I}$ and $\{x_{j}\}_{j \in \mathcal J}$ are linearly separable, and a linear classifier is trained by minimizing the cross entropy loss:
\begin{equation}
\min_{w, b} \ \sum\nolimits_{i \in \mathcal I} \log\left( { 1 + e^{-w^\top x_{i} - b}} \right) 
+\sum\nolimits_{j \in \mathcal J} \log\left( { 1 + e^{w^\top x_{j}+b}} \right), \nonumber
\end{equation}
via the gradient descent algorithm. Let $\langle \overline w, \cdot \rangle + B = 0$ denote the decision boundary obtained, 
and assume that $\overline w$ and $B$ are scaled such that
\[ \min_{i \in \mathcal I} \ \langle \overline w, x_{i} \rangle - \max_{j \in \mathcal J}\ \langle \overline w, x_{j} \rangle = 2.
\]
Define the set of indices for the support vectors as
\begin{align*} \mathcal I_\text{sup} & = \big\{
 i \in \mathcal I : \langle \overline w, x_{i} \rangle \le \langle \overline w, x_{i'} \rangle \ \forall {i'} \in \mathcal I \big\}, \\ \mathcal J_\text{sup} & = \big\{
 j \in \mathcal J : \langle \overline w, x_{j} \rangle \ge \langle \overline w, x_{j'} \rangle \ \forall {j'} \in \mathcal J \big\}. \end{align*}
If the support vectors  lie in an affine subspace, that is, if there exist a set of orthonormal vectors $\{r_k\}_{k \in K}$ and a set of scalars $\{\Delta_k\}_{k \in K}$ such that
\[ \langle r_k,  x_i \rangle =  \langle r_k, x_j \rangle = \Delta_k \quad \forall i \in \mathcal I_\text{sup}, \ \forall j \in \mathcal J_\text{sup}, \ \forall k \in K, \]
then the minimization of the cross-entropy loss yields a margin smaller than or equal to
\[ {1 \over \sqrt{ {1 \over {\gamma_\text{OPT}^{2}}} + B^2 \sum\nolimits_{k\in K}  \Delta_k^2  }} \]
where $\gamma_\text{OPT}$ denotes the optimal hard margin in the input space given by the SVM solution.
\label{theo:exact-subspace}
\end{theo}
\proof{
See Appendix \ref{app:theo-2}. \hfill \BlackBox
}

Theorem 2 points out that the margin of the classifier is determined by only the support vectors, which are the training points that are closest to the decision boundary of the classifier. The points that are further to the decision boundary have no effect on the margin, and there is a reason for this: their \emph{excitation} of the parameters do not \emph{persist} as the gradient descent algorithm continues to update the parameters. Once these points are correctly classified, their contribution to the gradient of the loss function is removed exponentially fast, and consequently, the dynamics of the algorithm is dominated by the other points.

Theorem 2 also shows that the margin obtained with cross-entropy minimization will be much smaller than the optimal value if the support vectors of the data set lie in a low dimensional affine subspace. Furthermore, as the dimension of this subspace decreases, the set $K$ in Theorem 2 grows in cardinality, and the margin of the classifier gets even smaller.

It is essential to note that the bias term $B$ in Theorem 2 is \emph{not} the bias of the training data set; it is the bias of the support vectors.
It is likely that the training data points have zero mean, whereas the support vectors have non-negligible bias.
In fact, the plot on the left of Figure 2 is intentionally chosen for this reason; it contains a data set with zero mean.

Note also that making the bias term $B$ zero requires the a priori knowledge of the set of support vectors --- which is not available until the algorithm has completed and the optimization problem is solved.
Therefore, the term $B$ can \emph{not} be made zero simply by preprocessing the data and removing its mean.
This is the primary reason why having a poor margin is unavoidable when the cross-entropy loss is used for low-dimensional data sets.


The statement of Theorem 2 requires that the support vectors lie exactly on an affine subspace. This condition is relaxed in Corollary 2.

\begin{corollary} Assume that the points $\{ x_{i}\}_{i \in \mathcal I}$ and $\{x_{j}\}_{j \in \mathcal J}$ in $\mathbb R^n$ are linearly separable and there exist a set of orthonormal vectors $\{r_k\}_{k \in K}$ and a set of scalars $\{\Delta_k\}_{k \in K}$ such that
\[ \langle r_k,  x_{i} \rangle \ge \Delta_k, \ \langle r_k, x_{j} \rangle \le \Delta_k \quad \forall i \in \mathcal I_\text{sup}, \ \forall j \in \mathcal J_\text{sup}, \ \forall k\in K.\]
Let $\langle \overline w, \cdot \rangle + B = 0$ denote the decision boundary obtained by minimizing the cross-entropy loss, as in Theorem \ref{theo:exact-subspace}. 
Then the minimization of the cross-entropy loss  yields  a margin smaller than or equal to
\[ {1 \over \sqrt{ B^2 \sum\nolimits_{k\in K}  \Delta_k^2  }}. \]
\end{corollary}

\proof{
See Appendix \ref{app:theo-2}. \hfill \BlackBox
}

Lastly, one could question why the weight parameters are not forced to be bounded in the analysis of this section by adding a penalty term into the training loss function in terms of some norm of the parameters. The short answer is that this section, along with all other parts of this work, aims to reveal how the richness of the training data arises within the dynamics of the gradient descent algorithm, and how it provides a guarantee for robustness. 
When the cost function is cross-entropy loss, there is no clear relationship between penalizing some norm of the weight parameters and improving the richness of the training data set, and this is why the inclusion of some regularization term, with its classical understanding, is avoided in the analysis of this section. This is clarified in Section \ref{section:richness} in detail.

\subsection{Input-Based Lipschitz Bounds for Two-Layer Neural Networks}

When a two-layer neural network is trained via the gradient descent algorithm, the dynamics of the algorithm will be nonlinear. If, in addition, the cost function used for training is the cross-entropy loss, the parameters of the network will grow unboundedly --- provided that there are enough parameters to classify every training point correctly with a proper initialization \citep{RechtUnderstanding}.
However, most analysis tools for nonlinear dynamical systems primarily focus on the behaviors of these systems around their equilibria, which are the fixed points of their dynamics, or their limiting behaviors inside bounded regions \citep{khalil1996nonlinear,sastry2013nonlinear}. 
Therefore, the common tools in nonlinear analysis will become inapplicable for the analysis of the cross-entropy loss. In this case, instead of studying the convergence of the actual values of the parameters, we can analyze the convergence of the ratio of the parameters. This leads us to consider an alternative concept of convergence: convergence in direction.

\begin{defn}\textbf{(Convergence in direction)} Given a set of functions, $W_k : [0, \infty) \mapsto \mathbb R^{m_k \times n_k}$ for $k \in [L]$, assume $\lim_{t \to \infty} \|W_k(t)\|_F = \infty$ for some $k \in [L]$. The set $(W_1, \dots, W_L)$ is said to converge in direction to $(\overline W_1, \dots, \overline W_L)$ if
\[ \lim_{t \to \infty} { {\|W_k(t) - h(t) \overline W_k \|}_F \over h(t)} = 0 \quad \forall k \in [L] \]
for some function $h: [0, \infty) \mapsto (0, \infty)$ diverging to $+\infty$, where $\overline W_k \in \mathbb R^{m_k \times n_k}$ is a matrix with bounded elements for each $k\in [L]$, and $\overline W_k \neq {\mathbf 0}$ at least for some $k \in [L]$.
\end{defn}

With this concept of convergence, we are ready to state the next theorem.

\begin{theo}
\label{theo:theo-3}
Assume that a two-layer neural network is trained to classify the sets $\{x_{i}\}_{i \in \mathcal I}$ and $\{x_{j}\}_{j\in \mathcal J}$ by minimizing the cross-entropy loss
\[  \sum\nolimits_{i \in \mathcal I} \log \left( 1 + e^{-w^\top {(Vx_{i} + b)}_+} \right) + 
\sum\nolimits_{j \in \mathcal J} \log \left( 1 + e^{w^\top {(Vx_{j} + b)}_+} \right)
\]
via the continuous-time gradient descent algorithm, where $w \in \mathbb R^r$, $V \in \mathbb R^{r \times n}$ and $b \in \mathbb R^r$. Assume all the training points are correctly classified\footnote{State-of-the-art neural networks are able to achieve zero training error even on randomly generated and randomly labeled data sets \citep{RechtUnderstanding}.}, and the parameters $(w, V, b)$ converge in direction to $(\overline w, \overline V, \overline b)$ as defined above.
Let $(\overline w, \overline V, \overline b)$ be scaled such that
\begin{subequations}
\label{eq:defn-support}
\begin{equation}
\label{eq:defn-support-a}
\overline  w ^\top (\overline V x_{i} + \overline b)_+ \ge 1 \quad \forall i \in \mathcal I, \end{equation}
\begin{equation}
\label{eq:defn-support-b}
\overline w^\top (\overline V x_{j} + \overline b)_+ \le -1 \quad \forall j \in \mathcal J \end{equation}
\end{subequations}
with either equality holding for at least one point. Let $\mathcal I_\text{sup}$ and $\mathcal J_\text{sup}$ denote the points that achieve equality in (\ref{eq:defn-support-a}) and (\ref{eq:defn-support-b}), respectively.
Then the Lipschitz constant of the mapping $x \mapsto \overline w^\top (\overline V x + \overline b)_+$ is upper bounded by
\[ {n_\text{\emph{node}}^\text{\emph{min}} \sqrt{n^\text{\emph{max}}_\text{\emph{sup}}} \over \sqrt{\underline \lambda}}, \]
where
\begin{itemize}
\item $n^\text{\emph{max}}_\text{\emph{sup}}$ is the maximum number of support vectors that activate the same hidden layer node:
\[ n^\text{\emph{max}}_\text{\emph{sup}} = \max_{k \in [r]} \ \max_{S \subseteq \mathcal I_\text{sup} \cup \mathcal J_\text{sup}} \ \left\{ |S| : \overline V_k x_s + \overline b_k > 0 \ \forall s \in S
\right\},
\]
\item $n_\text{\emph{node}}^\text{\emph{min}}$ is the minimum number of nodes that are activated by all of the support vectors:
\[ n_\text{\emph{node}}^\text{\emph{min}} = \min_{K \subseteq [r]} \left\{ |K| :   \max\nolimits_{k \in K} \overline V_kx_s + \overline b_k > 0 \ \forall s \in \mathcal I_\text{sup} \cup \mathcal J_\text{sup} 
\right\}, \]
\item $\underline \lambda$ is a lower bound for the minimum eigenvalue of $X_k^\top X_k$, where the \underline{columns} of $X_k$ are the support vectors activating node $k$.
\end{itemize}
\end{theo}

\proof{
See Appendix \ref{app:theo-3}. \hfill \BlackBox
}

Theorem 3 reveals a new notion of richness for the data set that is markedly different than the full-rankness of the data or of the support vectors; it is the linear independence of the support vectors. If the support vectors activating any one of the hidden-layer nodes are linearly dependent, then $\underline \lambda$ in Theorem 3 will be zero, and the bound will become void. Note that this notion of richness also reinforces the result of Theorem 2.

\section{Low-Dimensional Features on Multiple Layers}
\label{sec:low-dimensional}
As seen in Section 2 and Section 3, robustness guarantees for neural networks seem to require certain amount of richness in the training data set. While the full rankness of the training data activating each hidden-layer node provides persistency of excitation for a network trained with the squared-error loss, the use of cross-entropy loss necessitates the linear independence of all support vectors activating the same hidden-layer nodes. These two conditions are different, but both of them are likely to be violated if the training data is large in number and the features of the data are low-dimensional in some layer of the network.
Even though this appears to be a degenerate case, there are two main reasons why it is also the prevalent case:
\begin{enumerate}
\item The raw image, audio, and video data are low-dimensional by nature; this is the fundamental fact that has enabled data compression for decades by discovering and utilizing low-dimensional representations specific to different applications.
\item Even if the input data is full rank, the use of the gradient descent algorithm for training multiple-layer networks induces low-rank signals in the intermediate layers of the network. 
\end{enumerate}

The latter item has been observed in several empirical works, for example, in \citep{mahoney2018}. The following theorem confirms its validity for deep linear networks.

\begin{theo}
\label{prop:prop-1}
Assume that a deep linear network with $L$ layers is trained to classify the points in the sets $\{x_{i}\}_{i \in \mathcal I}$ and $\{x_{j}\}_{j \in \mathcal J}$ by minimizing 
\begin{equation}
\ell(W_1, \dots, W_L) =  \sum\nolimits_{s \in \mathcal I \cup \mathcal J} d(W_L \cdots W_1 x_s, y_s)  + \sum\nolimits_{k \in [L]}\mu_k \|W_k\|_F^2
\end{equation}
where $\{y_s : s \in \mathcal I \cup \mathcal J\}$ is the set of labels, $d(\cdot, \cdot)$ is any loss function differentiable in its first argument, $W_k \in \mathbb R^{n_k \times n_{k-1}}$ for all $k\in [L]$ and $n_L = 1$, i.e., the output of the network is scalar, and $\mu_k >0$ for all $k \in [L]$. If the gradient descent algorithm converges from a random initialization to a solution $(\overline W_1, \dots, \overline W_L)$, then each weight matrix $\overline W_k$ has rank 1, almost surely. 
\end{theo}

\proof{
 See Appendix \ref{app:prop-1}. \hfill \BlackBox
}

Theorem \ref{prop:prop-1} shows that adding the frobenious norms of the weight parameters, which is commonly referred to as adding weight-decay, causes all signals in the hidden layers to have rank 1 --- independently of the loss function used for training. The following theorem shows that when the cross-entropy loss is used, the result is the same even without the addition of weight-decay.

\begin{theo}
\label{app:prop-2}
Assume that the sets $\{x_{i}\}_{i \in \mathcal I}$ and $\{x_{j}\}_{j \in \mathcal J}$ are separable by a hyperplane passing through the origin, and a deep linear network with $L$ layers is trained to classify them by minimizing the cross-entropy loss:
\begin{equation}
\label{eqn:ce-loss}
\ell(W_1, \dots, W_L) = \sum\nolimits_{i \in \mathcal I} \log\left(1 + e^{W_L\cdots W_1 x_{i}} \right) + \sum\nolimits_{j \in \mathcal J} \log \left( 1 + e^{-W_L \cdots W_1 x_{j}} \right)
\end{equation}
via the continuous-time gradient descent algorithm:
\[ {dW_k(t) \over dt} = - {\partial \ell(W_1(t), \dots, W_L(t) ) \over \partial W_k(t)} \quad \forall k \in [L]. \]
If the weight matrices converge in direction to $(\overline W_1, \dots, \overline W_L)$ from a random initialization, then each $\overline{W}_k$ has rank 1 almost surely.
\end{theo}

\proof{
See Appendix \ref{appendix:prop-2}. \hfill \BlackBox
}

Theorem \ref{prop:prop-1} and Theorem \ref{app:prop-2} show that the conditions given in the previous sections for the persistency of excitation are easily violated when the model has multiple layers and the gradient descent algorithm is used for training.
This suggests that adding more training points into the data set will not be an effective approach to
provide persistency of excitation for the parameters of a neural network. The following section tries to address this issue by inserting exogenous disturbance to the dynamics of the gradient descent algorithm in order to ensure persistency of excitation for multiple-layer networks.

\section{Persistent Excitation of Parameters for Robustness}

It seems that multi-layer architectures make it challenging to ensure persistency of excitation, and this causes lack of robustness in multi-layer neural networks.
One straightforward way to remedy this problem
is to inject exogenous perturbations into the dynamics of the gradient descent algorithm so that even the parameters in the hidden layers receive persistent excitation during training.

In order to understand how this external excitation should be applied, 
this section will first look into 
the traditional procedure to introduce robustness for linear models, which is the addition of regularization term into the loss function, and describe how it could be reframed as a method to ensure persistent excitation of the model parameters. Then the second part of the section introduces an analogous procedure to persistently excite every parameter in a multiple-layer network.

\subsection{Reinterpreting Regularization as a Means to Ensure Persistency of Excitation}
\label{section:richness}

Given a set of points $\{x_i\}_{i \in \mathcal I}$ in $\mathbb R^n$ and corresponding target values $\{y_i\}_{i\in \mathcal I}$ in $\mathbb R$, consider the linear regression problem with the squared-error loss:
\begin{equation}
\label{prob:lin-regression}
\min_{w} \ \sum\nolimits_{i \in \mathcal I} ( w^\top x_i - y_i)^2. \end{equation}
 For this problem to have a unique solution, the set of training points $\{x_i\}_{i\in \mathcal I}$ needs to span the whole input space, $\mathbb R^n$. If the training data set is rank deficient, the usual remedy is to add a regularization term to the cost function \citep{hastie_09_elements-of.statistical-learning}:
\[ \min_{w} \ \sum\nolimits_{i \in \mathcal I} ( w^\top x_i - y_i )^2 + \mathcal R(w). \]
The regularization term $\mathcal R(w)$ is usually assumed to impose some prior knowledge about $w$, and it is chosen as a convex function of some norm of $w$, such as ${\|w\|}_2^2$ or ${\|w\|}_1$. This forces the parameters to be close to zero in some norm and restricts the class of functions that can be estimated by the model.

For the ridge regression problem corresponding to the choice $\mathcal R(w) = \lambda \|w\|_2^2$ for any $\lambda >0$:
\begin{equation} \label{eqn:ridge} \min_{w} \ \sum\nolimits_{i \in \mathcal I} ( w^\top x_i - y_i )^2 + \lambda \|w\|_2^2, \end{equation}
it is an ordinary observation that adding the regularization term is equivalent to adding $n$ ghost measurements, since we have
\[ \lambda \|w\|_2^2 = \sum\nolimits_{i=1}^n \left(w^\top \big(\sqrt{\lambda} e_i\big) - 0 \right)^2, \]
where $e_i \in \mathbb R^n$ is the $i$-{th} standard basis vector of $\mathbb R^n$. Alternatively, an equivalent problem could also be formulated as
\begin{equation}
\label{ridge:expectation}
 \min_{w} \ \mathbb E_{\{\xi_i\}_{i \in \mathcal I}} \sum\nolimits_{i \in \mathcal I} \left( w^\top (x_i + \xi_i) - y_i\right)^2 \end{equation}
where $\xi_i \in \mathbb R^n$ is any random vector with 
\[\mathbb E [\xi_i] = 0, \quad \mathbb E [ \xi_i \xi_i^\top ] = {\lambda \over | \mathcal I |}I\]
for each $i \in \mathcal I$. In particular, the probability distributions for $\{\xi_i\}_{i\in \mathcal I}$ could be chosen to be discrete for practical purposes, and the ridge regression could be considered as augmenting the training data set by perturbing each training point in, for example, $2n$ different directions and creating $2n$ new training points, and consequently, producing a new training data set with $2n|\mathcal I|$ points in total.

The equivalence of problems (\ref{eqn:ridge}) and  (\ref{ridge:expectation}) is critical: even though the common regularization terms in use are originally introduced to impose a prior knowledge about the weight parameters and to bound the class of estimated functions, they could have also been introduced as a way to create a larger data set from the original training points and solve the optimization problem with this larger data set. This equivalence seems to depend on the choice of squared $\ell_2$ norm as the regularization term, but the following theorem suggests another way to augment the data set for linear regression, which preserves the equivalence for a much larger class of regularization terms.

\begin{theo}
\label{theo:theo-4}
Given a set of points $\{x_i\}_{i\in \mathcal I}$ in $\mathbb R^n$ and corresponding target values $\{y_i\}_{i \in \mathcal I}$ in $\mathbb R$, consider the following two problems:
\begin{subequations}
\begin{equation}
    \label{prob-1}
    \min_{w} \  \sum_{i \in \mathcal I} (y_i - w^\top x_i)^2 + \lambda \|w\|_p^m,
\end{equation}
\begin{equation}
    \label{prob-2}
    \min_{w} \  \sum_{i \in \mathcal I} {1 \over 2} \left( y_i - \min_{d : \|d\|_q \le \epsilon} w^\top (x_i + d) \right)^2 + {1 \over 2}
    \left( y_i - \max_{d: \|d\|_q \le \epsilon} w^\top (x_i + d) \right)^2,
\end{equation}
\end{subequations}
where $\|\cdot \|_p$ and $\|\cdot \|_q$ are dual norms, $m \in [1,\infty)$ is some fixed number, and $\lambda, \epsilon \in (0, \infty)$ are hyperparameters\footnote{The case for higher-dimensional target values is stated in Corollary \ref{corol:corol-3}.}. For each $\lambda$, there exists some $\epsilon$ such that the solutions of the two problems are identical. Conversely, for each $\epsilon$, there exists some $\lambda$ such that the solutions of the two problems are identical.
\end{theo}
\proof{
See Appendix \ref{app:theo-4}. \hfill \BlackBox
}

Note that problem (\ref{prob-2}) solves the regression problem with the two extreme points on the $q$-norm ball around each training point, where the radius of the ball is given by $\epsilon$. This provides a way to ensure that every point inside the ball around $x_i$ is assigned a value close to $y_i$ at the end of optimization --- thereby indirectly solving the regression problem with a richer data set while avoiding the generation of a large set of points as in (\ref{ridge:expectation}). 

Theorem 6 shows that the following are equivalent \textbf{for linear regression with squared-error loss}:
\begin{enumerate}
    \item bounding or penalizing any norm of the weight parameters,
    \item augmenting the training data set by inflating the training points and solving the regression problem with this larger data set,
    \item inducing generalization, i.e., providing comparable results on the training data and on unseen points that are close to the training data.
\end{enumerate}
It is clear that items 2 and 3 are equivalent irrespective of the loss function used for regression. Equivalence of item 1 to the others, however, does depend on the loss function. To see this, consider the cross-entropy loss with two sets of points $\{x_{i}\}_{i \in \mathcal I}$ and $\{x_{j}\}_{j \in \mathcal J}$:
\[ \min_{w} \sum_{i \in \mathcal I} \log \Big( 1 + e^{-w^\top x_{i}} \Big) + \sum_{j \in \mathcal J} \log\Big(1 + e^{w^\top x_{j}}\Big), \]
and assume that these sets are linearly separable. 
If we inflate the training data as in~problem (\ref{prob-2}), we obtain
\begin{gather}
\label{robust-logistic}
\min_{w} \  \bigg\{  {1\over 2} \sum_{i \in \mathcal I} \log\Big(1 + e^{-w^\top x_{i} -\epsilon \|w\|_q} \Big) + \log\Big(1 + e^{-w^\top x_{i} + \epsilon \|w\|_q}\Big)  \\
\qquad + {1 \over 2}  \sum_{j\in \mathcal J} \log\Big(1 + e^{w^\top x_{j} + \epsilon \|w\|_q}\Big) + \log\Big(1 + e^{w^\top x_{j} - \epsilon \|w\|_q}\Big) \bigg\} \nonumber
\end{gather}
where $\epsilon$ is the $p$-norm of the disturbance applied to each training point. This problem is not convex, but more importantly, this problem does not attain its global minimum at a finite point --- as long as $\epsilon$ does not exceed the maximum $p$-norm margin of the original data set. This is because the two sets remain linearly separable even after each point is inflated by $\epsilon$ amount in $p$-norm.  Consequently, the normal vector of any hyperplane separating these inflated points gives a direction for $w$ to drive the cost function arbitrarily close to zero by increasing $\|w\|_q$. Note that the infimum of the cost function is zero, and any bounded vector will be strictly suboptimal.

Problem (\ref{robust-logistic}) gives an example of the case where the optimal solution is unbounded but generalizes perfectly for the unseen data near the training points. Moreover, it shows that enforcing boundedness of the parameters could prevent the discovery of the optimal solution.
This simple example reveals why \textbf{the generalization of the performance of a model does not necessarily require, or entail, the boundedness of the model parameters}. This fact is the main reason why the analyses in Section \ref{section:cross-entropy} do not involve any penalty term to impose a bound on the model parameters.

Lastly, it should be noted that problem (\ref{prob-2}) is different from adversarial training \citep{madry}. Adversarial training regards the loss function used in the empirical risk minimization as the ultimate goal and tries to minimize this loss function under worst disturbances:
\[ \min_{w} \ \sum_{i \in \mathcal I} \max_{d : \|d\|_p \le \epsilon} \ell(w, x_i + d).
\]
This overlooks the fact that the loss function is only a proxy for the actual task. 
Problem (\ref{prob-2}), on the other hand, tries to
enforce that the estimate values be close to the target values for all neighbors of the training data --- which is the very property needed for robustness. 
In the next subsection, the formulation of problem (\ref{prob-2}) will form our basis for generating persistently exciting perturbations for neural networks.

\subsection{A Training Scheme for Persistent Excitation of Neural Network Parameters}

Problem (\ref{prob-2}) shows that for each training point, perturbing the input of a linear mapping in two directions such that the output of the mapping is maximized and minimized provides persistency of excitation for the parameters of the model. An analogous way to excite the parameters of a multiple-layer neural network could be inserting perturbation to every signal preceding an affine operation inside the network.
To illustrate, consider an $L$-layer feedforward network:
\begin{subequations}
\label{L-layer-network}
\begin{align}
    h_0(x) & = x, \\
    h_{j}(x) & = \sigma_{j}(W_j h_{j-1}(x) + b_j) \quad j=1, 2, \dots, L,
\end{align}
\end{subequations}
where $\{W_j\}_{j\in [L]}$ and $\{b_j\}_{j\in [L]}$ are the weight and bias parameters of the network, and $\{\sigma_j(\cdot)\}_{j\in [L]}$ are the nonlinear activation functions. The parameter $W_j$ corresponds to a linear operation in the \hbox{$j$-th} layer of the network, such as a matrix multiplication or a convolution. We can insert perturbations to this network as
\begin{subequations}
\label{tilde-h-network}
\begin{align}
    \tilde h_0(x;d) & = x,\\
    \tilde h_{j}(x;d) &= \sigma_j(W_j [\tilde h_{j-1}(x;d) + d_j] + b_j) \quad j=1,2, \dots, L,
\end{align}
\end{subequations}
where $d = (d_1, d_2, \dots, d_{L})$ is the set of perturbations applied to each layer of the network.
Then, solving the regression problem with the cost function
\[
\sum_{i \in \mathcal I} {1 \over 2}\left( y_i - \min_{d \in \mathcal D} \tilde h_L(x_i; d) \right)^2 + {1 \over 2} \left( y_i - \max_{d \in \mathcal D}\tilde h_L(x_i; d) \right)^2,
\]
where $\mathcal D$ is the allowed set of perturbations, should force the output of the network to remain close to the target values despite moderate changes in the input and in the signals in the hidden layers.

It might seem curious that we perturb the hidden layers as well only to guarantee robustness against perturbations in the input of the network. To reiterate the necessity for this, reconsider the neural network in (\ref{L-layer-network}) as composition of two mappings:
\[ h_L(x) = (h_{L \gets m} \circ h_m)(x) \]
where $L$ is the total number of layers with $L \ge 3$, $m$ is the depth of an intermediate layer satisfying $2 \le m \le L-1$, and $h_{L\gets m}$ represents the mapping from the output of the $m$-th layer of the network to the output of the last layer. As we have discussed in Section 4, assume that the training data set $\{x_i\}_{i \in \mathcal I}$ lies in a low dimensional subspace within the input space. Without loss of generality, let
\[ \langle x_i, e_1 \rangle = \langle x_i, e_2 \rangle =  0 \quad \forall i \in \mathcal I, \]
where $e_1$ and $e_2$ are the first and the second standard basis vectors of $\mathbb R^n$. Assume hypothetically~that the training procedure causes one of the nodes in the $m$-th layer to output the product of the first and the second coordinates of the input points. Then for all the training points, this node will output zero, and how the output of this node is propagated in $h_{L \gets m}$ will not be observed at all. Now assume that in an attempt to provide robustness, we only perturb the input points but not the signals in the intermediate layers. If the allowed perturbations are bounded by $\epsilon$ in $\ell_\infty$ norm, the output of the node of interest will be in the range $[-\epsilon^2, \epsilon^2]$, and this will likely not suffice to observe the effect of this node on the output of the network. Consequently, even if the output of the network is predicted well for all the training data and the points in their $\epsilon$-neighborhood, the reaction of the network to the points not in the close neighborhood of the training data will be little known. This argument is well supported by the observations in recent works, which show that training a neural network by (adversarially) perturbing the training data provides a large margin for the training data, but not for the unseen test data \citep{madry}.

Before proceeding to the next section, we summarize in Algorithm 1 the procedure for training a neural network while ensuring persistent excitation of the parameters.
For simplicity, the algorithm is described for the stochastic gradient method with momentum; a batch of training points can also be used at every step to compute the gradient update.

\begin{algorithm}
\caption{Training with Persistent Excitation}
\label{alg:myalgo}
\begin{algorithmic}[1]
\State \textbf{input:} training data $\{x_i\}_{i \in \mathcal I}$, \newline \indent \quad target values $\{y_i\}_{i \in \mathcal I}$, \newline \indent \quad 
neural network $f_\theta(x; d) \equiv \tilde h_L(x; d)$ in (\ref{tilde-h-network}), \newline 
\indent \quad set of allowed perturbations $\mathcal{D}$, \newline \indent \quad learning rate $\eta$, \newline \indent \quad momentum $\gamma$
\State \textbf{output:} network parameters $\theta$
\State \textbf{initialize:} $\Delta \theta \gets 0$
\Repeat
   \State randomly choose $i \in \mathcal I$
   \State $d_1 \gets \argmax_{d \in \mathcal D} f_\theta(x_i; d)$
   \State $d_2 \gets \argmin_{d \in \mathcal D} f_\theta(x_i; d)$
   \State $\Delta \theta \gets \gamma \Delta \theta + (1 - \gamma) \nabla_{\theta} \left[ (f_\theta(x_i;d_1) - y_i)^2 + (f_\theta(x_i;d_2) - y_i)^2
   \right]$
   \State $\theta \gets \theta - \eta \Delta \theta$
\Until{training is complete}
\end{algorithmic}
\end{algorithm}

\section{Experimental Results}

In this section, we test Algorithm 1 on a binary classification task. Only two classes of images, the horses and the planes, have been chosen from the CIFAR-10 data set for the classification task. The same network architecture is used in all of the experiments: two convolutional layers followed by two fully-connected layers with leaky-ReLu activations. Neither batch-normalization nor drop-out is implemented in the experiments. The reason why we test Algorithm 1 on a binary classification task and how it could be extented to multiple-class classification tasks are discussed in Section \ref{sec:discussion}.

The first experiment is to demonstrate that classical training paradigm with the squared-error loss and the cross-entropy loss is not effective in providing a large margin for the unseen data points. A four-layer network is trained with these loss functions with no exogenous perturbation for persistency of excitation.
The plots in Figure 4 show the percentage of points the network misclassify versus the amount of disturbance needed in the input of the network to cause the misclassification, which is computed by using the projected gradient attack algorithm \citep{madry, foolbox}. In this sense, the plots are analogous to the cumulative distribution function of the margin of the data; and the lower the plot, the more robust the network. It is observed that neither the squared-error loss nor the cross-entropy loss is effective in providing a large  margin for the unseen test data.

\begin{figure}[ht]
    \centering
    \includegraphics[width=0.49\textwidth]{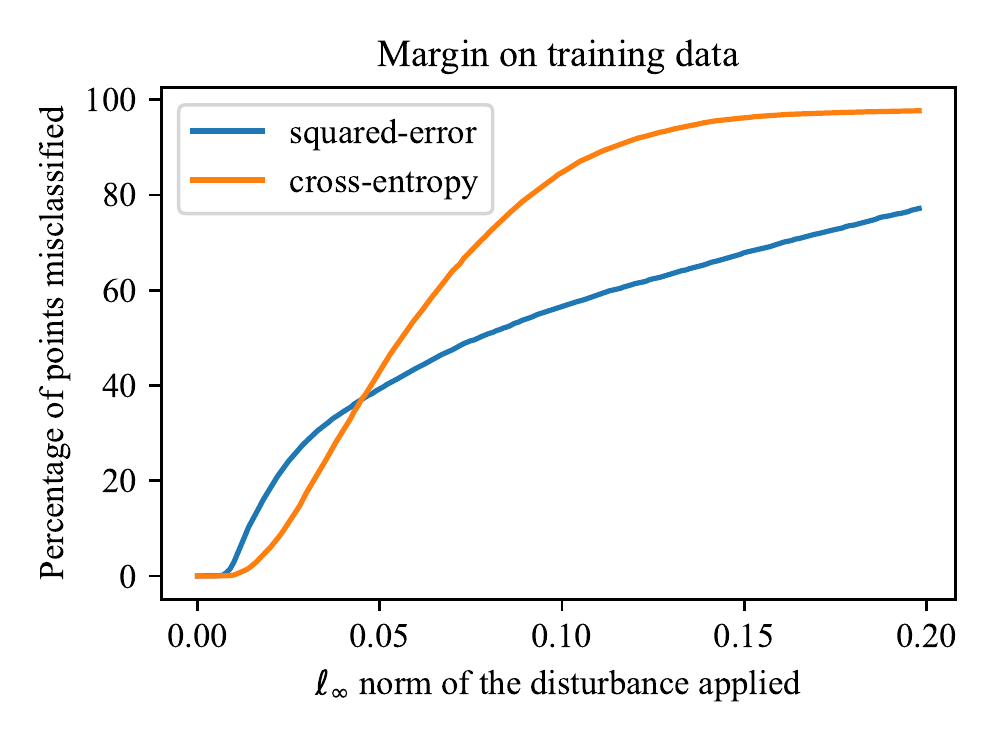}
    \includegraphics[width=0.49\textwidth]{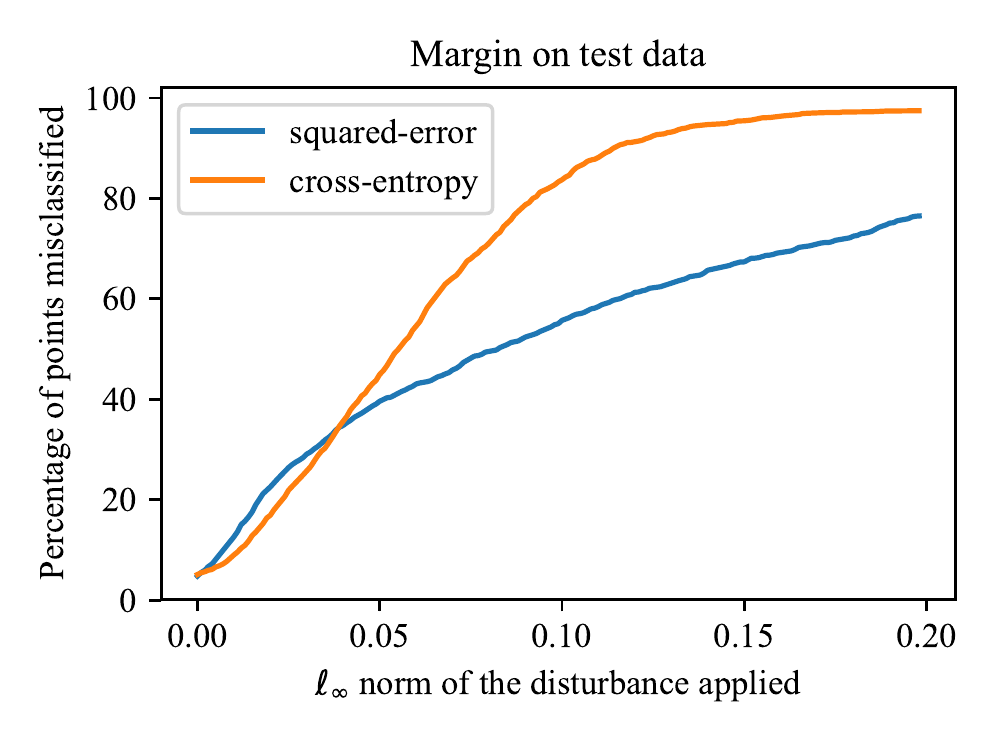}
    \caption{A four-layer network is trained for a binary-classification task with the cross-entropy~and the squared-error loss. The plots demonstrate the percentage of points the network misclassifies versus the amount of disturbance needed in the input of the network to cause the misclassification. The curves are analogous to the cumulative distribution function of the margin between the decision boundary of the classifier and the data set. The lower the curve, the more robust the network. The plots show that neither the squared-error loss nor the cross-entropy loss is effective in providing a large margin for the unseen test data. Note that a perturbation of magnitude 1.0 changes a white pixel to an exact gray.
    }
    \label{fig:both-poor-margin}
\end{figure}

The next experiment is to see the effectiveness of Algorithm 1. We train the same four-layer network with Algorithm 1 by using three different sets of perturbation, which are $\ell_\infty$ balls with radii 0.005, 0.010 and 0.020 on all layers. Comparing the left plot of Figure~4 with the left plot of Figure~5, we observe that applying a small persistently-exciting perturbation increases the margin of the training data conspicuously. We also observe that the training error is not able to reach zero when the perturbation is relative large, as shown by the curve for 0.020 perturbation. Nevertheless, the same magnitude of perturbation attains the largest margin on the test data.

\begin{figure}[ht]
    \centering
    \includegraphics[width=0.49\textwidth]{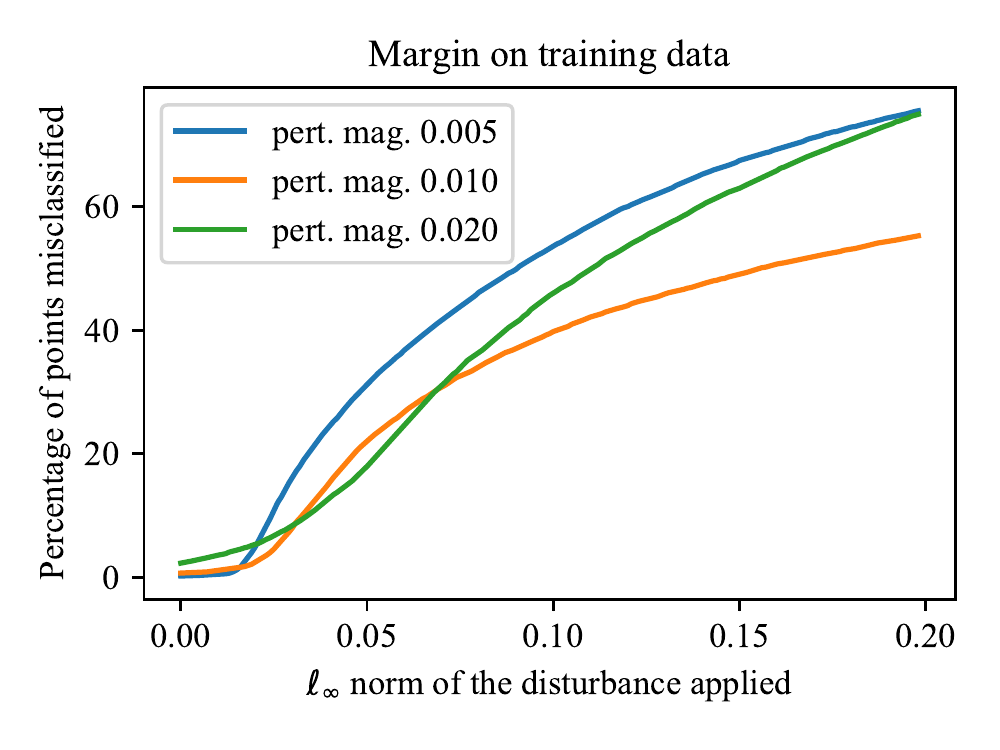}
    \includegraphics[width=0.49\textwidth]{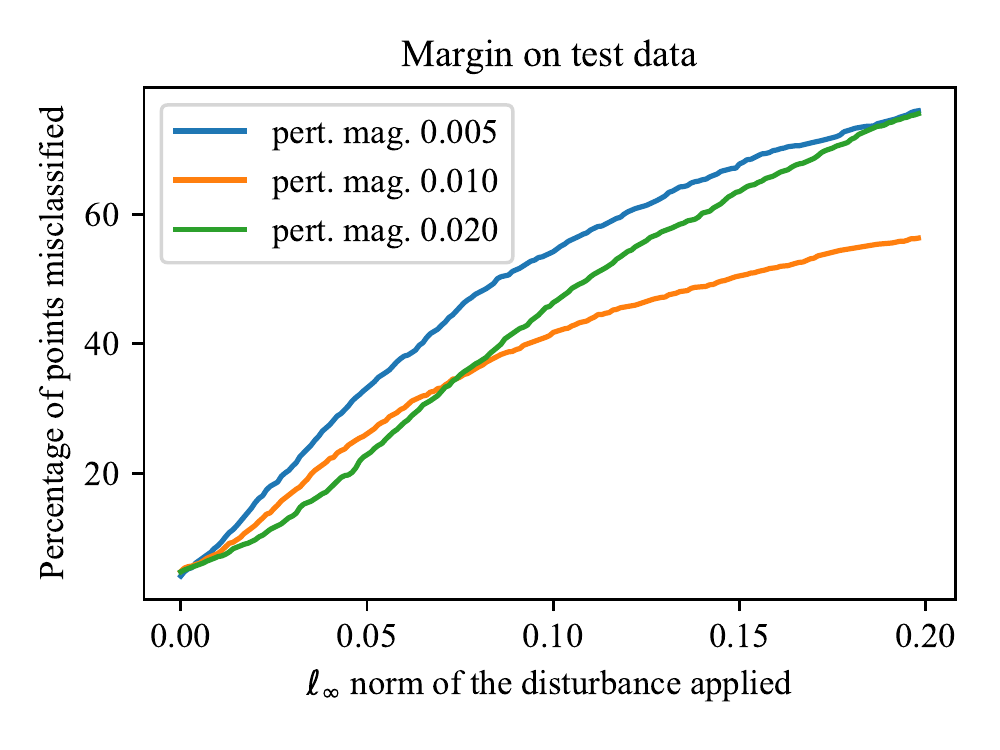}
    \caption{The effect of training with persistent excitation as described in Algorithm 1. Small perturbations for persistent excitation improves the margin for the training data. The larger pertubations prevent the training error from reaching zero; however, they yield larger margins on the test data.}
    \label{fig:algo-1}
\end{figure}

The last experiment is to demonstrate the necessity of perturbing all layers, and not only the first layer, in order to ensure persistent excitation of the parameters of the network. Figure 6 shows the margin distribution of the same network trained in two different ways: perturbing all layers of the network as described in Algorithm 1 and perturbing only the input of the network during training, similar to adversarial training \citep{madry}. For both cases, the perturbations are restricted to be in $\ell_\infty$ ball with radius 0.020. We observe that perturbing only the first layer of the network during training substantially increases the margin of the training data; however, this does not correspond to an improvement for the margin of the test data. In contrast, when all layers are perturbed for persistency of excitation as described in Algorithm 1, the margin of the training data becomes a good indicator of the margin of the test data.

\begin{figure}[ht]
    \centering
    \includegraphics[width=0.5\textwidth]{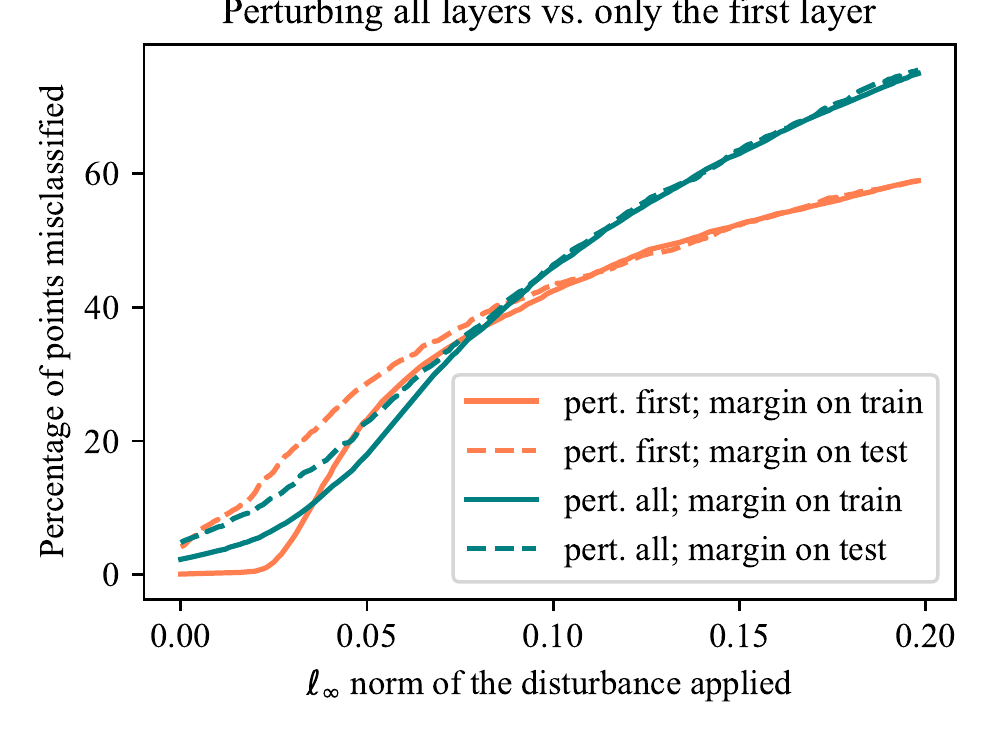}
    \caption{The same network is trained in two different ways: by perturbing every layer of the network as described in Algorithm 1 for persistency of excitation, and by perturbing only the training data (the first layer) similar to adversarial training. Perturbing only the first layer substantially increases the margin for the training data; however, this is not reflected in the test data. In contrast, when all layers are perturbed for persistency of excitation as outlined in Algorithm~1, the margin of the training data becomes a good indicator of the margin of the test data.}
    \label{fig:all-vs-first}
\end{figure}

\section{Discussion}
\label{sec:discussion}

In this section, we discuss why we have restricted our analyses mostly to binary classification, how it could be extended to multiple-class classification, how Algorithm 1 could be applied to different loss functions, and how its performance could be improved. We also compare our key results with the previous works.

\subsection{Emphasis on Binary Classification and Extensions to Multi-Class Classification}
In this work, when studying the robustness of neural networks, we have focused on binary classification tasks. There are two main reasons for this.

\begin{enumerate}
    \item Consider the subject of hypothesis testing. The analysis of binary hypothesis testing naturally precedes the study of multiple hypotheses, and it is essential to reveal the critical factors and the bottlenecks in hypothesis testing, such as the trade-off between Type-I and Type-II errors, the ROC curve, and the Neyman-Pearson rule \citep{poor2013introduction,keener2011theoretical}. Similarly, the study of robustness of machine learning models must start with the simplest case where robustness can be analyzed clearly if we truly want to understand what causes lack of robustness and how it can be addressed. This simplest case appears to be the binary classification.

    \item Binary classification is used in many critical tasks. For example, given the medical data of a patient, the decision of whether the patient has a certain disease or not is a binary classification task. Similarly, in computer security, the decision of whether a certain face or fingerprint belongs to the set of authorized users is a binary classification task.
\end{enumerate}

There are two straighforward ways to extend Algorithm 1 to the classification of multiple classes.
The first approach is to use a hierarchy of binary classifiers. Given $m$ different classes, each training point can be assigned to a class with at most $\lceil \log_2(m) \rceil$ binary classifiers. The performance of this classifier depends on the choice of ordering and the groups of classes.

The second approach avoids using a hierarchy of classifiers and tries to impose fairness by employing equidistant target values inside a higher dimensional space. This approach is based on the following corollary to Theorem \ref{theo:theo-4}.
\begin{corollary}
\label{corol:corol-3}
Given a set of points $\{x_i\}_{i \in \mathcal I}$ in $\mathbb R^n$ and corresponding target values $\{y_i\}_{i\in \mathcal I}$ in $\mathbb R^m$, consider the following two problems:
\begin{gather}
    \min_{W} \quad  \sum_{i \in \mathcal I} {\left\| y_i - W x_i \right\|}_2^2 + \lambda {\|W\|}_{p_1, p_2}^2
    \end{gather}
    and
    \begin{subequations}
    \begin{align}
    \min_{W} & \quad \sum_{i \in \mathcal I} {\left\| y_i - W(x_i + d_{i})
    \right\|}_2^2 + {\left\| y_i - W(x_i - d_{i})
    \right\|}_2^2\\
    \subjectto & \quad d_{i} \in \argmax_{d : \|d\|_{p_1} \le \epsilon} \,  \max_{v : \|v\|_{q_2} \le 1} \ \left\{ v^\top
        W (x_i + d) - v^\top Wx_i \right\}
         \quad \forall i \in  \mathcal I, \label{eq:corol-3-b}
\end{align}
\end{subequations}
where ${\|\cdot\|}_{p_2}$ and ${\|\cdot\|}_{q_2}$ are dual norms, and ${\|W\|}_{p_1, p_2}$ denotes the induced norm of $W$:
\[ {\|W\|}_{p_1, p_2} = \max_{u: \|u\|_{p_1}=1} \|Wu\|_{p_2}. \]
 For each $\lambda$, there exists some $\epsilon$ such that the solutions of the two problems are identical. Conversely, for each $\epsilon$, there exists some $\lambda$ such that the solutions of the two problems are identical. \hfill \BlackBox
\end{corollary}

The involvement of $x_i$ in (\ref{eq:corol-3-b}) might seem unnecessary since the model is linear, but remember that we introduce this corollary to find persistently exciting perturbations for nonlinear classifiers.
To demonstrate the use of Corollary 3 for a multi-class classifier in an example, assume that the training data $\{x_i\}_{i \in \mathcal I}$ in $\mathbb R^n$ come from four different classes, and the labels for these classes are $\{y_i\}_{i \in \mathcal I}$. 
For these labels, we can associate four equidistant vectors $z^{(1)}, z^{(2)}, z^{(3)}, z^{(4)} \in \mathbb R^3$ such that
\begin{align*} y_i \in \{z^{(1)}, z^{(2)}, z^{(3)}, z^{(4)}\} & \quad \forall i \in \mathcal I,\\
 {\|z^{(i)} - z^{(j)}\|}_2 = 1 \quad \  & \quad \forall i,j \in \{1, 2, 3, 4\}, \ i \neq j.\end{align*}
Then the problem of training a classifier $f_\theta : \mathbb R^n \to \mathbb R^3$ with squared-error loss can be formulated~as
\[ \min_\theta \sum\nolimits_{i \in \mathcal I} \left\| f_\theta(x_i) - y_i \right\|_2^2. \]
Corresponding to point $x_i$, we can find two different persistently-exciting perturbations, $d_{i_1}$ and $d_{i_2}$, by solving
\[ 
v_i = \argmax_{v: {\|v\|}_2 \le 1} \, \max_{d \in \mathcal D} \ \left\{ v^\top f_\theta(x_i; d) - v^\top f_\theta(x_i) \right\},
\]
\[d_{i_1} \in \argmax_{d \in \mathcal D}  \ v_i^\top f_\theta(x_i; d), \]
\[
d_{i_2} \in \argmin_{d \in \mathcal D} \ v_i^\top f_\theta(x_i; d),
\]
where $f_\theta(x;d)$ is the network represented by $\tilde h_L(x; d)$ in (\ref{tilde-h-network}). 
With these perturbations, we can write the optimization problem with persistent excitation as
\[ \min_\theta \sum\nolimits_{i \in \mathcal I} 
\| f_\theta(x_i;d_{i_1}) - y_i \|_2^2
+ 
\| f_\theta(x_i;d_{i_2}) - y_i \|_2^2. \]
Given the solution $\hat \theta$ to this problem, we can finally describe the classifier as
\[ x \mapsto \argmin\nolimits_{z \in \{z^{(1)}, z^{(2)}, z^{(3)}, z^{(4)} \}} \| f_{\hat \theta}(x) - z\|_2. \]

\subsection{Alternative Loss Functions and Perturbation Sets}

The proposed algorithm for persistent excitation considers only the squared-error loss function. An analogous approach for the cross-entropy loss is possible, but the perturbations for the hidden layers need to be either not additive or not bounded. This is because the cross-entropy loss yields unbounded growth of the parameters of the network, and consequently, the signals in the hidden layers will become unboundedly large. In this case, the addition of finite perturbations will have no effect on the output of the network. One potential alternative is introducing multiplicative perturbations to the signals in the hidden layers.

In Algorithm 1, the set of allowed perturbations $\mathcal D$ is left as an arbitrary set. In our experiments, this set is chosen to be an $\ell_\infty$ ball with the same radius for all layers.
Applying perturbations with different magnitudes to different layers of the network could improve the robustness of the network, while preserving its overall accuracy.

\subsection{Related Works}
There are three main subjects that are closely related to our results.

\textbf{1. Logistic regression and maximum margin:} Recently there has been a line of work \citep{Soudry-March-2018, ji2018risk} claiming that linear classifiers trained with  the gradient descent algorithm achieve the maximum margin if the cross-entropy loss function is used --- which sounds contrary to the results in Section 3. There are two reasons for this discrepancy.
The first reason is an unfortunate misnomer: support vector machines with no bias term are also called the maximum-margin solution in some sources, e.g. \citep{rosset2004margin, shalev2014understanding}, even though the decision boundary found with these problems is forced to pass through the origin, and hence, necessarily suboptimal.
The second and the more critical reason is that the analysis in \citep{Soudry-March-2018} starts with removing the necessity for a bias term by inserting the training data into an augmented space; consequently, all claims about maximum margin are stated for this augmented state. In other words,
\textbf{this so-called maximum margin is attained in an augmented space; not in the original space the training points lie in}.  
However, as demonstrated in Figure~2, the true margin between the training data and the decision boundary of the classifier could be substantially smaller than the optimal margin that could be obtained in the original space.
This distinction has not been emphasized, nor mentioned, in \citep{Soudry-March-2018}, and the ambiguity has led to the misperception that the gradient descent algorithm on the cross-entropy loss yields the \emph{true} maximum-margin solution in the original space --- despite the fact that neural networks are consistently shown to misclassify inputs that are barely different than their \emph{training} data \citep{Goodfellow2015Adversarial}, and therefore, there is no possibility they could have a large margin between their decision boundary and the training data.
Unfortunately, this misconception has been continued and even been used in an attempt to explain the generalization of neural networks \citep{wei2019on}.

\textbf{2. Importance of low-dimensionality:} As discussed in Section 4, most of the data used~for training neural networks, such as image, audio or video, lie in a low-dimensional space in some domain. Furthermore, the iterative training algorithms induce low-dimensional signals in the intermediate layers of neural networks independently of the training data. This fact has been critical to draw the conclusion that it is not possible to achieve robustness in neural networks simply by retrieving more training data. If this low-dimensionality is overlooked, the conclusion will be different.
Indeed, \cite{schmidt2018} analyzes the robustness of classifiers with full-rank, non-degenerate data sets and single-layer, linear classifiers, and arrives at the conclusion that the robustness could be achieved with more training data.

\textbf{3. Trade-off between training speed and generalization:}
It has been shown in recent works that adaptive gradient algorithms do not usually yield solutions which generalize well, even though they accelerate the training procedure substantially \citep{wilson2017marginal}.
The concept of persistent excitation gives an insight about this: the adaptive gradient methods remove the effect of correctly classified points too quickly and do not allow these points to persistently excite the parameters of the model. The same argument applies to the training loss functions: if a loss function removes the effect of correctly classified points too quickly, then it could yield a worse generalization, as seen in the case of cross-entropy loss.

\section{Conclusion}

In this work, we looked into the dynamics of the gradient descent algorithm
to understand why neural networks lack robustness and how this vulnerability could be addressed.
We identified sufficient conditions on the training data sets to ensure robustness of two-layer neural networks.
We showed that for classification tasks, multi-layer networks are unlikely to satisfy these conditions, and the signals interacting with the parameter estimates consequently fail to remain persistently exciting.
This lack of persistent excitation removes all the guarantees about how the network will react to inputs that are different than its training data.

To provide a solution for this problem, we studied the classical regularization terms used for linear models closely and reframed them in the context of persistency of excitation.
Based~on~this framework, we introduced a new algorithm to generate persistently exciting perturbations and tested this algorithm on a binary classification task.
We demonstrated that the proposed algorithm improves the correspondence between the margins of the training data and the test data.

\bibliography{references}

\appendix



\section{Proof of Theorem \ref{theo:theo-1}}
\label{app:theo-1}
To begin with, assume $b$ is fixed and not updated by the gradient descent algorithm. Let $(\hat W, \hat V)$ denote the local optimum that the algorithm has converged. For point $x_i$, let $G_i \in \{0,1\}^{r\times r}$ be the diagonal matrix that satisfies
\[ \hat W(\hat Vx_i + b)_+ = \hat WG_i(\hat Vx_i + b). \]
The update rule of the gradient descent algorithm is given as
\begin{subequations}
\label{sys:grad}
\begin{align}
 W  & \gets W - \delta \left\{
\sum\nolimits_{i \in \mathcal I} \left[ W G_i (V x_i + b) - f(x_i) \right] (Vx_i + b)^\top G_i^\top
\right\}, \\
 V & \gets V - \delta \left\{
\sum\nolimits_{i \in \mathcal I} G_i^\top W^\top \left[ W G_i(V x_i + b) - f(x_i) \right] x_i^\top
\right\}, \end{align}
\end{subequations}
followed by
\begin{align*}
(G_i)_{kk} & \gets \mathbf{1}_{\{ V_kx_i + b_k > 0\}} \quad \forall k \in [r], \ \forall i \in \mathcal I,
\end{align*}
where $(G_i)_{kk}$ denotes the $k$-th diagonal element of the matrix $G_i$, and $V_k$ and $b_k$ represent the $k$-th rows of $V$ and $b$. If the algorithm has converged, switching between 1 and 0 must have ended in a finite time. Therefore, around an equilibrium point, we will assume $G_i$ is fixed for all $i \in \mathcal I$.

Like all dynamical systems, the gradient descent algorithm (\ref{sys:grad}) can converge to an equilibrium from randomly chosen close neighbors of this point only if the equilibrium is \emph{stable in the sense of Lyapunov} \citep{sastry2013nonlinear}. A necessary condition for an equilibrium of a nonlinear dynamical~system to be stable is that the linear approximation of this dynamical system not be unstable around the same equilibrium. We will use this fact to obtain a necessary condition for the algorithm to converge to an equilibrium from any randomly chosen initial point.

Linearization of the system (\ref{sys:grad}) around the equilibrium $(\hat W, \hat V)$ gives
\begin{subequations}
\label{system:linearized}
\begin{align}
(W-\hat W) & \gets (W-\hat W) -f_1(W-\hat W) - f_2(V-\hat V), \\
(V -\hat V) & \gets (V-\hat V ) - f_3(W -\hat W) - f_4(V - \hat V),
\end{align}
\end{subequations}
where
\begin{align*}
f_1(\Delta W) & = \delta \sum_{i \in \mathcal I} \Delta  W G_i(\hat V x_i + b)(\hat V x_i + b)^\top G_i^\top, \\
f_2(\Delta V) & = \delta \sum_{i \in \mathcal I} \hat WG_i \Delta V x_i x_i^\top \hat V G_i^\top + \delta \sum_{i \in \mathcal I} [\hat W G_i(\hat V x_i + b) - f(x_i)]x_i^\top\Delta  V^\top G_i^\top, \\
f_3(\Delta W) & = \delta \sum_{i \in \mathcal I} G_i^\top \hat W^\top\Delta  W G_i(\hat V x_i + b)x_i^\top + \delta \sum_{i \in \mathcal I} G_i^\top \Delta W^\top [\hat W G_i(\hat V x_i + b) - f(x_i)]x_i^\top, \\
f_4(\Delta V) & = \delta \sum_{i \in \mathcal I} G_i^\top \hat W^\top \hat W G_i \Delta V x_i x_i^\top.
\end{align*}
We have the equality
\[ \langle \Delta W, f_2(\Delta V) \rangle = \langle f_3(\Delta W), \Delta V \rangle \quad \forall \Delta W \in \mathbb R^{m\times r}, \ \forall \Delta V \in \mathbb R^{r \times n}; \]
therefore, the linearized system (\ref{system:linearized}) can be represented by a symmetric matrix\footnote{This is also a direct consequence of the fact that $f_2$ and $f_3$ are the Jacobians of the gradient of the training loss function with respect to the same parameters in different orders.}, and the system (\ref{system:linearized}) is stable in the sense of Lyapunov only if the matrix corresponding to the system
\begin{subequations}
\begin{align}
\label{system2}
(W-\hat W) & \gets (W-\hat W) -f_1(W-\hat W) \\
(V-\hat V) & \gets (V-\hat V ) - f_4(V-\hat V)
\end{align}
\end{subequations}
has all of its eigenvalues inside the unit circle. This implies that the convergence of the system (\ref{system:linearized}) requires the largest eigenvalues of the mappings $f_1(\cdot)$ and $f_4(\cdot)$ to be smaller than $2$.

The largest eigenvalue of the mapping $f_1(\cdot)$ being smaller than 2 implies that
\[ e_k^\top \sum\nolimits_{i \in \mathcal I} G_i(\hat Vx_i + b)(\hat V x_i + b)^\top G_i^\top e_k \le {2 \over \delta} \quad \forall k\in [r],\]
where $e_k$ denotes the standard basis vector with 1 in its $k$-th element. Then,
\[  e_k^\top \sum\nolimits_{i \in \mathcal I_k}(\hat V x_i + b)(\hat V x_i + b)^\top  e_k  \le {2 \over \delta} \quad \forall k\in [r]\]
where $\mathcal I_k$ denotes the set of indices of the points that activate node $k$ at equilibrium, i.e., 
 \[  \mathcal I_k = \{ i\in \mathcal I : e_k^\top (\hat V x_i + b) > 0 \}. \]
 Let $\hat V_k$ and $b_k$ denote the $k$-th rows of $\hat V$ and $b$. Then we need
 \[ \sum\nolimits_{i \in \mathcal I_k} (\hat V_k x_i + b_k)(x_i^\top \hat V_k^\top + b_k) \le {2 \over \delta}, \]
 or equivalently,
 \[ \hat V_k^\top \left( \sum\nolimits_{i\in \mathcal I_k} x_i x_i^\top \right) \hat V_k + 2 b_k \hat V_k^\top \left(\sum\nolimits_{i\in \mathcal I_k} x_i \right)+ b_k^2  - {2 \over \delta} \le 0. \]
 This is a quadratic inequality in $\hat V_k$, and the largest value $\big\| \hat V_k\big\|_2$ can take is upper bounded by the larger root of
 \[ \lambda_\text{min} \left( \sum\nolimits_{i \in \mathcal I_k} x_i x_i^\top \right) {\big\|\hat V_k\big\|}_2^2 - 2 \left|b_k\right| \left\| \sum\nolimits_{i \in \mathcal I_k} x_i \right\|_2 {\big\|\hat V_k \big\|}_2  + b_k^2 - {2 \over \delta} = 0.
 \]
 Therefore, we have
 \[ {\big\| \hat V_k \big\|}_2 \le { \left| b_k \right| \mu_k \over \lambda_k^\text{min} } + {1 \over \lambda^\text{min}_k } \sqrt{ b_k^2 {(\mu_k)}^2 + \lambda_k^\text{min} \left( {2 \over \delta} - b_k^2 \right) }
 \]
 where
 \[ \mu_k = \left\| \sum\nolimits_{i \in \mathcal I_k} x_i \right\|_2, \]
 \[ \lambda^\text{min}_k = \lambda_\text{min} \left( \sum\nolimits_{i \in \mathcal I_k} x_i x_i^\top \right). \]
 As an upper bound independent of $k$, we can write
 \[ {\big\|\hat V_k \big\|}_2 \le { \overline \mu \|b\|_\infty \over \underline \lambda } +  \sqrt{ {\overline \mu^2 \|b\|_\infty^2 \over \underline \lambda^2}  +  { 1\over \underline \lambda}\left( {2 \over \delta} - \|b\|_\infty^2 \right) }  \qquad \forall k\in[r]\]
 where $\overline \mu = \max_{k \in [r]} \mu_k$ and $ \underline \lambda  = \min_{k \in [r]} \lambda^\text{min}_k$.
 
 So far we have only used the fact that the largest eigenvalue of $f_1(\cdot)$ is less than 2. Similarly, the largest eigenvalue of $f_4(\cdot)$ being less than 2 implies that
 \[  \sum\nolimits_{i \in I_k}  e_k^\top \hat W^\top \hat W e_k  \cdot \lambda_\text{min} \left( \sum\nolimits_{i \in \mathcal I_k} x_ix_i^\top \right) \le {2 \over \delta}. 
 \]
 If $\hat W_k$ denotes the $k$-th column of $\hat W$, we have
 \[ {\big\| \hat W_k \big\|}_2 \le \sqrt{{ 2 \over \delta \underline \lambda} } \qquad \forall k\in [r]. \]
 
 Given the estimates $\hat W$ and $\hat V$, for every $x\in \mathbb R^n$, the function estimated by the network is $ \hat W(\hat V x + b)_+$, and the Lipschitz constant of this estimate is bounded by
 \[  \sum\nolimits_{k=1}^r \left\| \mathbf{1}_{ \left\{ \hat V_kx + b > 0 \right\}}\hat W_k \hat V_k \right\|_F, \]
which is further bounded by
\[ n_\text{active}^\text{max} \left(\sqrt{{2 \over \delta \underline \lambda}} \right) \left(
{ \overline \mu \|b\|_\infty \over \underline \lambda } +  \sqrt{ {\overline \mu^2 \|b\|_\infty^2 \over \underline \lambda^2}  +  { 1\over \underline \lambda}\left( {2 \over \delta} - \|b\|_\infty^2 \right) } 
\right)
\]
and
\[ n_\text{active}^\text{max} \left( \sqrt{2 \over \delta \underline \lambda} \right) \left( {2 \overline \mu \|b\|_\infty \over \underline \lambda} + \sqrt{{1 \over \underline \lambda}\left| {2 \over \delta} - \|b\|_\infty^2\right| }
\right)
\]
where $n_\text{active}^\text{max}$ is the maximum number of nodes that a point in $\mathbb R^n$ can activate, i.e., 
\[ n_\text{active}^\text{max} = \max_{x \in \mathbb R^n} \sum\nolimits_{k=1}^r \mathbf{1}_{\left\{ \hat V_k x + b > 0 \right\}}. \]

To complete the proof, now assume that $b$ is not fixed and it is also updated by the gradient descent algorithm.
We can write the linearization of the update rule for $(W,V,b)$ as
\begin{subequations}
\label{sys:last}
\begin{align}
    \begin{bmatrix}
     W \\ V^\top
    \end{bmatrix} & \gets \begin{bmatrix}
     W \\ V^\top
    \end{bmatrix} -  g_1\left( \begin{bmatrix}
     W \\ V^\top
    \end{bmatrix} \right) - g_2\left( b \right),  \\
    b & \gets b - g_3\left( \begin{bmatrix}
     W \\ V^\top
    \end{bmatrix} \right) - g_4\left(b \right),
\end{align}
\end{subequations}
where $g_1, g_2, g_3$ and $g_4$ are the linear operators obtained by taking Jacobians of the gradients of the training loss function with respect to $W$, $V$ and $b$.
Similar to $f_2$ and $f_3$ in system (\ref{system:linearized}), the operators $g_2$ and $g_3$ are the Hermitian of each other, and therefore, the matrix corresponding to the system (\ref{sys:last}) is still symmetric. As a result, its eigenvalues are less than 1 in magnitude only if its diagonal sub-blocks have eigenvalues less than 1 in magnitude, which leads to the identical condition for the case with fixed~$b$.  \hfill $\blacksquare$

\section{Proof of Theorem \ref{theo:theo-2}}
\label{app:theo-2}

\begin{lemma}[Adapted from Theorem 3 of \citealt{Soudry-March-2018}] Given two sets of points $\{x_i: i \in \mathcal I\}$ and $\{x_j: j \in \mathcal J\}$ that are linearly separable in $\mathbb R^n$, let $\tilde x_i$ and $\tilde x_j$ denote $[ x_i^\top \ 1]^\top$ and $[x_j^\top \ 1]^\top$, respectively, for all $i \in \mathcal I$, $j \in \mathcal J$. Then the iterate of the gradient descent algorithm, $\tilde w(t)$, on the cross-entropy loss function
\begin{equation*}  
\min_{\tilde w \in \mathbb R^{n+1}} \sum\nolimits_{i \in \mathcal I} \log(1 + e^{-\tilde w^\top \tilde x_i}) + \sum\nolimits_{j \in \mathcal J} \log(1 + e^{\tilde w^\top \tilde x_j}) \label{loss_cross_ent} \end{equation*}
 with a sufficiently small step size will converge in direction:
 \[ \lim_{t \to \infty} {\tilde w(t) \over \|\tilde w(t)\|} = {\overline w \over \|\overline w\|}, \] 
where $\overline w$ is the solution to 
\begin{align}
\label{eqn:svm-in-difference}
\underset{z \in \mathbb R^{n+1}}{\text{\emph{minimize}}}  &\quad  \|z\|^2  \\
\text{\emph{subject to}} & \quad \langle z, \tilde x_i \rangle \ge 1 \quad \forall i \in \mathcal I, \nonumber \\
&\quad  \langle z, \tilde x_j\rangle  \le -1 \quad  \forall j \in \mathcal J. \nonumber \end{align}
\end{lemma}

\noindent {\bf Proof of Theorem \ref{theo:theo-2}.} 
Assume that $\overline w = u + \sum_{k \in K} \alpha_k r_k$, where $u \in \mathbb R^{n}$ and $\langle u, r_k \rangle = 0$ for all $k \in K$. By denoting $z = [w^\top \ b]^\top$, 
the Lagrangian of the problem (\ref{eqn:svm-in-difference}) can be written as
\begin{gather*} {1\over 2} \| w\|^2 + {1 \over 2}b^2 + \sum\nolimits_{i\in \mathcal I} \mu_i ( 1 - \langle  w,  x_i \rangle - b) + \sum\nolimits_{j \in \mathcal J} \nu_j (-1 + \langle  w,  x_j \rangle +b ), \end{gather*}
where $\mu_i \ge 0$ for all $i \in \mathcal I$ and $\nu_j \ge 0$ for all $j \in \mathcal J$.
KKT conditions for the optimality of $\overline w$ and $B$ requires that
\[ \overline w = \sum_{i\in \mathcal I} \mu_i  x_i - \sum_{j \in \mathcal J} \nu_j  x_j,\ \
 B = \sum_{i \in \mathcal I} \mu_i - \sum_{j \in \mathcal J} \nu_j,\]
and consequently, for each $k \in K$,
\begin{eqnarray*} \langle \overline w, r_k \rangle & = & \sum\nolimits_{i\in \mathcal I} \mu_i \langle  x_i , r_k \rangle - \sum\nolimits_{j\in \mathcal J} \nu_j \langle x_j, r_k \rangle \\ & = & \sum\nolimits_{i \in \mathcal I} \Delta_k\mu_i - \sum\nolimits_{j \in \mathcal J} \Delta_k \nu_j = B\Delta_k. \end{eqnarray*}
Then, we can write $\overline w$ as
\[ \overline w = u + \sum\nolimits_{k\in K} B\Delta_k r_k. \]
Let $ \langle w_\text{SVM}, \cdot \rangle + b_\text{SVM} = 0$ denote the hyperplane obtained as the solution of maximum hard-margin SVM in the original space --- not in the augmented space. Then $w_\text{SVM}$ solves
\begin{align}  \underset{w}{\text{{minimize}}} & \quad \|  w \|^2  \label{eq:new-svm} \\
\text{subject to} & \quad \langle  w, x_i - x_j\rangle \ge 2 \quad \forall i \in \mathcal I, \forall j \in \mathcal  J. \nonumber
\end{align}
Since the vector $u$ also satisfies $\langle u, x_i - x_j \rangle = \langle  w,  x_i -  x_j \rangle \ge 2$ for all $i \in \mathcal I,j \in \mathcal J$, we have $\|u\| \ge \|w_\text{SVM}\| = {1 \over \gamma_\text{OPT}}$. As a result, the margin obtained by minimizing the cross-entropy loss is 
\[ {1 \over \|\overline w\|} = { 1 \over \sqrt{ \|u\|^2 + \sum  \|B\Delta_k r_k\|^2}} \le {1 \over \sqrt{ {1 \over \gamma^2_\text{OPT}} + B^2\sum \Delta_k^2}}. \]
\hfill $\blacksquare$

\noindent {\bf Proof of Corollary 2.}
If $B <0$, we could consider the hyperplane $\langle \overline w, \cdot \rangle - B = 0$ for the points $\{- x_i\}_{i\in \mathcal I}$ and $\{-x_j\}_{j \in \mathcal J}$, which would have the identical margin due to symmetry. Therefore, without loss of generality, assume $B \ge 0$. As in the proof of Theorem 2, KKT conditions for the optimality of $\overline w$ and $B$ requires
\[ \overline w = \sum_{i \in \mathcal I} \mu_i  x_i - \sum_{j \in \mathcal  J} \nu_j  x_j , \ \ B = \sum_{i \in \mathcal I} \mu_i - \sum_{j \in \mathcal J} \nu_j \]
where $\mu_i \ge 0$ and $\nu_j \ge 0$ for all $i \in \mathcal  I, j \in \mathcal J$. Note that for each $k \in K$,
\begin{eqnarray*}
 \langle \overline w, r_k \rangle & = & \sum\nolimits_{i \in \mathcal I} \mu_i \langle  x_i, r_k\rangle - \sum\nolimits_{j \in \mathcal J} \nu_j \langle x_j, r_k\rangle \\
 & = & B\Delta_k + \sum\nolimits_{i \in \mathcal I} \mu_i (\langle x_i, r_k\rangle - \Delta_k ) \\
 & & - \sum\nolimits_{j \in \mathcal J} \nu_j (\langle - x_j, r_k\rangle - \Delta_k ) \ \ge\  B\Delta_k.
 \end{eqnarray*}
 Since $\{r_k\}_{k\in K}$ is an orthonormal set of vectors,
 \[ \|\overline w\|^2 \ge \sum\nolimits_{k\in K} \left\langle \overline w, r_k \right\rangle^2 \ge \sum\nolimits_{k \in K} B^2 \Delta_k^2.\]
 The result follows from the fact that ${\|\overline w\|}^{-1}$ is an upper bound on the margin. \hfill $\blacksquare$

\section{Proof of Theorem \ref{theo:theo-3}}
\label{app:theo-3}
\begin{lemma} 
\label{lemma-for-lemma}
The direction parameters $\overline w, \overline V$ and $\overline b$ satisfy
\[ \|\overline w\|_2^2 = \|\overline V \|_F^2 + \|\overline b\|_2^2. \]
\end{lemma}

\noindent
\textbf{Proof} The continuous-time gradient descent algorithm gives
\begin{subequations}
\label{eqn:continuous-dynamics}
\begin{align}
{dw \over dt} & = \sum_{i \in \mathcal I} G_{i}(V x_{i} + b) { e^{-w^\top G_{i}(Vx_{i} + b)} \over 1 +  e^{-w^\top G_{i}(Vx_{i} + b)}}
- \sum_{j \in \mathcal J} G_{j}(V x_{j} + b) { e^{w^\top G_{j}(Vx_{j} + b)} \over 1 +  e^{w^\top G_{j}(Vx_{j} + b)}} \\
{dV \over dt} & = \sum_{i \in \mathcal I} G_{i} w x_{i}^\top { e^{-w^\top G_{i}(Vx_{i} + b)} \over 1 +  e^{-w^\top G_{i}(Vx_{i} + b)}} 
- \sum_{j \in \mathcal J} G_{j} w x_{j}^\top { e^{w^\top G_{j}(Vx_{j} + b)} \over 1 +  e^{w^\top G_{j}(Vx_{j} + b)}} \\
{db \over dt} & = \sum_{i \in \mathcal I} G_{i} w { e^{-w^\top G_{i}(Vx_{i} + b)} \over 1 +  e^{-w^\top G_{i}(Vx_{i} + b)}} 
- \sum_{j \in \mathcal J} G_{j} w  { e^{w^\top G_{j}(Vx_{j} + b)} \over 1 +  e^{w^\top G_{j}(Vx_{j} + b)}}
\end{align}
\end{subequations}
Note that, for all $t \ge 0$ we have
\[ {d \over dt} \langle w(t), w(t) \rangle = {d \over dt} \langle V(t), V(t) \rangle + {d \over dt} \langle b(t), b(t) \rangle. \]
Consequently, for all $t \ge 0$:
\[ { \|w(t)\|_2^2 - \|w(0)\|_2^2 \over h^2(t)} = 
{\|V(t)\|_F^2 - \|V(0)\|_F^2 \over h^2(t)} + {\|b(t)\|_2^2 - \|b(0)\|_2^2 \over h^2(t) }, \]
which proves that
\begin{equation*} \|\overline w\|_2^2 = \|\overline V\|_F^2 + \|\overline b\|_2^2.  \end{equation*}
\hfill $\blacksquare$

\begin{lemma} The solution obtained by the continuous-time gradient descent algorithm satisfies
\begin{align*}
\overline w & = \sum_{i \in \mathcal I} \mu_{i} G_{i} (\overline V x_{i} + \overline b) 
- \sum_{j \in \mathcal J} \mu_{j} G_{j} (\overline V x_{j} + \overline b)  \\
\overline V & = \sum_{i \in \mathcal I} \mu_{i} G_{i} \overline w x_{i}^\top 
- \sum_{j \in \mathcal J} \mu_{j} G_{j} \overline w x_{j}^\top \\
\overline b & = \sum_{i \in \mathcal I} \mu_{i} G_{i} \overline w 
- \sum_{j \in \mathcal J} \mu_{j} G_{j} \overline w
\end{align*}
for some set of nonnegative scalars $\{\mu_s : s \in \mathcal I \cup \mathcal J\}$. 
\end{lemma}

\noindent
\textbf{Proof} Given the dynamics (\ref{eqn:continuous-dynamics}) of the gradient descent algorithm, define the set of support vectors as
\begin{align*}
\mathcal I^\text{sup} & = \left\{ i \in \mathcal I : \lim_{t \to \infty} {e^{-w^\top(Vx_{i'} + b)_+} \over e^{-w^\top (Vx_i + b)_+}} < \infty \ \forall i' \in \mathcal I, \ 
\lim_{t \to \infty} {e^{w^\top(Vx_{j'} + b)_+} \over e^{-w^\top (Vx_i + b)_+}} < \infty \ \forall j' \in \mathcal J
 \right\}, \\
 \mathcal J^\text{sup} & = \left\{ j \in \mathcal J : \lim_{t \to \infty} {e^{-w^\top(Vx_{i'} + b)_+} \over e^{w^\top (Vx_j + b)_+}} < \infty \ \forall i' \in \mathcal I, \ 
\lim_{t \to \infty} {e^{w^\top(Vx_{j'} + b)_+} \over e^{w^\top (Vx_j + b)_+}} < \infty \ \forall j' \in \mathcal J
 \right\}.
\end{align*}
Note that the points corresponding to these indices dominate the dynamics of the algorithm as the training continues. Let $\overline b_{s_1}, \overline b_{s_2}$ be two nonzero coordinates of $\overline b$, let $x_S$ be any of the support vectors. Then,

\begin{align*} 
{ \overline b_{s_1} \over \overline b_{s_2}} & = \lim_{t \to \infty} {b_{s_1}(t) \over b_{s_2}(t)} \\
&  = \lim_{t \to \infty} { {d{ b}_{s_1} \over dt} \over {d{ b}_{s_2}\over dt} } \\
& = \lim_{t \to \infty} {
\sum_{i \in \mathcal I} e_{s_1}^\top G_{i} w { e^{-w^\top G_{i}(Vx_{i} + b)} \over 1 +  e^{-w^\top G_{i}(Vx_{i} + b)}} 
- \sum_{j \in \mathcal J} e_{s_1}^\top G_{j} w  { e^{w^\top G_{j}(Vx_{j} + b)} \over 1 +  e^{w^\top G_{j}(Vx_{j} + b)}}
\over
\sum_{i \in \mathcal I} e_{s_2}^\top G_{i} w { e^{-w^\top G_{i}(Vx_{i} + b)} \over 1 +  e^{-w^\top G_{i}(Vx_{i} + b)}} 
- \sum_{j \in \mathcal J} e_{s_2}^\top G_{j} w  { e^{w^\top G_{j}(Vx_{j} + b)} \over 1 +  e^{w^\top G_{j}(Vx_{j} + b)}}
} \\
& = \lim_{t \to \infty} {
\sum_{i \in \mathcal I} e_{s_1}^\top G_{i} w { e^{-w^\top G_{i}(Vx_{i} + b)} \over  e^{w^\top G_{S}(Vx_{S} + b)}} 
- \sum_{j \in \mathcal J} e_{s_1}^\top G_{j} w  { e^{w^\top G_{j}(Vx_{j} + b)} \over e^{w^\top G_{S}(Vx_{S} + b)}} 
\over
\sum_{i \in \mathcal I} e_{s_2}^\top G_{i} w { e^{-w^\top G_{i}(Vx_{i} + b)} \over e^{w^\top G_{S}(Vx_{S} + b)}} 
- \sum_{j \in \mathcal J} e_{s_2}^\top G_{j} w  { e^{w^\top G_{j}(Vx_{j} + b)} \over e^{w^\top G_{S}(Vx_{S} + b)}} 
} \\
& =
{ \sum_{i \in \mathcal I} e_{s_1}^\top G_{i} \overline w  \mu_{i} - \sum_{j \in \mathcal J} e_{s_1}^\top G_{j} \overline w  \mu_{j}
\over
\sum_{i \in \mathcal I} e_{s_2}^\top G_{i} \overline w  \mu_{i} - \sum_{j \in \mathcal J} e_{s_2}^\top G_{j} \overline w  \mu_{j}
}
\end{align*}
where
\begin{align*}
  \mu_{i} & \propto \lim_{t \to \infty} { e^{-w^\top G_{i}(Vx_{i} + b)} \over  e^{w^\top G_{S}(Vx_{S} + b)}} \quad \forall i \in \mathcal I, \\
 \mu_{j} & \propto \lim_{t \to \infty} { e^{-w^\top G_{j}(Vx_{j} + b)} \over  e^{w^\top G_{S}(Vx_{S} + b)}} \quad \forall j \in \mathcal J.
 \end{align*}

Similarly, if $e_{s_1}^\top \overline b$ and $e_{s_2}^\top \overline V e_{s_3}$ are two nonzero elements of $\overline b$ and $\overline V$, then
\begin{align*}
{ e_{s_1}^\top \overline b \over e_{s_2}^\top \overline V e_{s_3} }
 & =  \lim_{t \to \infty} {e_{s_1}^\top b \over e_{s_2}^\top V e_{s_3}} \\
  & =  \lim_{t \to \infty} {e_{s_1}^\top {d b \over dt} \over e_{s_2}^\top {d V \over dt} e_{s_3}} \\
  & = \lim_{t \to \infty} {
\sum_{i \in \mathcal I} e_{s_1}^\top G_{i} w { e^{-w^\top G_{i}(Vx_{i} + b)} \over  e^{w^\top G_{S}(Vx_{S} + b)}} 
- \sum_{j \in \mathcal J} e_{s_1}^\top G_{j} w  { e^{w^\top G_{j}(Vx_{j} + b)} \over e^{w^\top G_{S}(Vx_{S} + b)}} 
\over
\sum_{i \in \mathcal I} e_{s_2}^\top G_{i} w x_{i} e_{s_3} { e^{-w^\top G_{i}(Vx_{i} + b)} \over e^{w^\top G_{S}(Vx_{S} + b)}} 
- \sum_{j \in \mathcal J} e_{s_2}^\top G_{j} w x_{j} e_{s_3}  { e^{w^\top G_{j}(Vx_{j} + b)} \over e^{w^\top G_{S}(Vx_{S} + b)}} 
} \\
& =
{ \sum_{i \in \mathcal I} e_{s_1}^\top G_{i} \overline w  \mu_{i} - \sum_{j \in \mathcal J} e_{s_1}^\top G_{j} \overline w  \mu_{j}
\over
\sum_{i \in \mathcal I} e_{s_2}^\top G_{i} \overline w x_{i} e_{s_3} \mu_{i} - \sum_{j \in \mathcal J} e_{s_2}^\top G_{j} \overline w x_{j} e_{s_3}  \mu_{j}
}.
\end{align*}
As a result, we have
\begin{align*}
\overline V & = \sum_{i \in \mathcal I} \mu_{i} G_{i} \overline w x_{i}^\top 
- \sum_{j \in \mathcal J} \mu_{j} G_{j} \overline w x_{j}^\top \\
\overline b & = \sum_{i \in \mathcal I} \mu_{i} G_{i} \overline w 
- \sum_{j \in \mathcal J} \mu_{j} G_{j} \overline w
\end{align*}
for some set of nonnegative scalars $\{ \mu_s : s \in \mathcal I \cup \mathcal J \}$.

An identical analysis of $\overline w_{s_1} / \overline w_{s_2}$ for two nonzero coordinates of $\overline w$ shows that
\[ \overline w  = \sum_{i \in \mathcal I} \alpha \mu_{i} G_{i} (\overline V x_{i} + \overline b) 
- \sum_{j \in \mathcal J} \alpha \mu_{j} G_{j} (\overline V x_{j} + \overline b) \]
for some $\alpha \in (0, \infty)$. In order to find the value of $\alpha$, note that
\begin{align*}
\langle \overline w, \overline w \rangle
& =  \left\langle \overline w, \sum_{i \in \mathcal I} \alpha \mu_{i} G_{i} (\overline V x_{i} + \overline b) 
- \sum_{j \in \mathcal J} \alpha \mu_{j} G_{j} (\overline V x_{j} + \overline b) \right\rangle \\
& = \alpha \left\langle \sum_{i \in \mathcal I}  \mu_{i}G_{i} \overline w x_{i}^\top - \hspace{-1mm}
 \sum_{j \in \mathcal J}  \mu_{j}G_{j} \overline w x_{j}^\top , \overline V \right\rangle 
 + \alpha \left\langle \sum_{i \in \mathcal I}  \mu_{i}G_{i} \overline w - \hspace{-1mm} \sum_{j \in \mathcal J}  \mu_{j} G_{j} \overline w, \overline b \right\rangle \\
 & = \alpha \langle \overline V, \overline V \rangle  + \alpha \langle \overline b, \overline b \rangle
 \end{align*}
which shows that $\alpha$ must be 1 due to Lemma \ref{lemma-for-lemma}. This completes the proof.
\hfill $\blacksquare$

\noindent\textbf{Proof of Theorem \ref{theo:theo-3}.} For any $k \in [r]$, let $\overline w_k$, $\overline V_k$ and $\overline b_k$ denote the $k$-th row of $\overline w$, $\overline V$ and $\overline b$, respectively. From Lemma \ref{lemma-for-lemma}, we have that
\[ \overline w_k^2 = \| \overline V_k\|_2^2 + \overline b_k^2. \]
Let $\mathcal I^k \subseteq \mathcal I$ and $\mathcal J^k \subseteq \mathcal J$ denote the support vectors that activate the $j$-{th} node in the hidden layer, i.e., 
\begin{align*}
\mathcal I^k & = \left\{ i \in \mathcal I : \overline V_kx_i + \overline b_k > 0, \ \overline w^\top (\overline V x_i + \overline b)_+ = 1 \right\}, \\
\mathcal J^k & = \left\{ j \in \mathcal J : \overline V_kx_j + \overline b_k > 0, \ \overline w^\top (\overline V x_j + \overline b)_+ = -1 \right\}.
\end{align*}
Then we have
\begin{align*}
\overline w_k & = \sum\nolimits_{i \in \mathcal I^k} \mu_{i}(\overline V_k x_{i} + \overline b_k) - \sum\nolimits_{j \in \mathcal J^k} \mu_{j} (\overline V_k x_{j} + \overline b_k) \\
\overline V_k & = \sum\nolimits_{i \in \mathcal I^k} \mu_{i}\overline w_k x_{i}^\top - \sum\nolimits_{j \in \mathcal J^k} \mu_{j} \overline w_k x_{j}^\top \\
\overline b_k & = \sum\nolimits_{i \in \mathcal I^k} \mu_{i} \overline w_k - \sum\nolimits_{j \in \mathcal J^k} \mu_{j} \overline w_k.
\end{align*}
Plugging the expressions for $\overline V_k$ and $\overline b_k$ into that of $\overline w_k$: 
\[ \overline w_k = \overline w_k \mu^\top X^\top_k X_k \mu + {1 \over \overline w_k} \overline b_k^2,\]
which gives
\begin{align*} \mu^\top X^\top_k X_k \mu = 1 - {\overline b_k^2 \over \overline w_k^2} & \implies  \lambda_\text{min}(X^\top_kX_k)  \|\mu\|_2^2 \le 1 - {\overline b_k^2 \over \overline w_k^2}, \\
& \implies
\lambda_\text{min}(X^\top_k X_k)  {\|\mu\|_1^2 \over n_k} \le 1 - {\overline b_k^2 \over \overline w_k^2}, \end{align*}
where $X_k$ is a matrix with columns $\{x_{i} : i \in \mathcal I^k \}$ and $\{-x_{j} : j \in \mathcal J^k\}$, $\mu$ is a column vector with elements $\{ \mu_s : s \in \mathcal I^k \cup \mathcal J^k \}$, and $n_k$ is the number of support vectors that activate node $k$. Note that if the columns of $X_k$ are not linearly independent, then $\mu_s$ can be arbitrarily large for any support vector $x_s$. From the last inequality, we have
\[ \sum\nolimits_{s \in \mathcal I^k \cup \mathcal J^k} \mu_s  \le {\sqrt{ |\mathcal I^k| + |\mathcal J^k|} \over \sqrt{\lambda_\text{min}(X^\top_k X_k)}}. \]

From the definition of the support vectors, we also have that
\begin{align*}
\sum\nolimits_{k \in [r]} \overline w_k^2 & = \sum\nolimits_{k \in [r]} \left( \sum\nolimits_{i \in \mathcal I^k} \mu_{i}\overline w_k(\overline V_k x_{i} + \overline b_k) - \sum\nolimits_{j \in \mathcal J^k} \mu_{j} \overline w_k(\overline V_k x_{j} + \overline b_k) \right) \\
& =  \sum\nolimits_{s \in \mathcal I \cup \mathcal J} \mu_s
\end{align*}
due to complementary slackness condition.

Our goal is to bound the Lipschitz constant of the estimate $\hat f(x) = \overline w^\top (\overline V x + \overline b)_+$: 
\begin{align*}
 \sum\nolimits_{k \in [r]} |\overline w_k| \|\overline V_k\|_2  & \le \sum\nolimits_{k \in [r]} \overline w_k^2 \\
 & = \sum\nolimits_{s \in \mathcal I \cup \mathcal J} \mu_s \\
 & \le {n^\text{min}_\text{node} \sqrt{n^\text{max}_\text{sup}} \over \min_{k \in [r]} \sqrt{ \lambda_\text{min}(X^\top_k X_k)}}
\end{align*}
where $n^\text{min}_\text{node}$ denotes the minimum number of nodes that are activated by all of the support vectors:
\begin{equation*} n^\text{min}_\text{node} = \min_{K \subseteq [r]} \left\{ |K| :  \forall s \in \mathcal I^\text{sup} \cup \mathcal J^\text{sup}   \ \exists k \in K \text{ such that } \overline V_kx_s + \overline b_k > 0
\right\}. \end{equation*}
\hfill $\blacksquare$

\section{Proof of Theorem \ref{prop:prop-1}}
\label{app:prop-1}

The first order stationarity condition implies that
\[ {\partial \ell \over \partial W_k}  = 0 \quad \forall k \in [L],\]
which yields
\begin{align*} 
0 & = \sum\nolimits_{s \in \mathcal I \cup \mathcal J} \overline W_{k+1}^\top \cdots \overline W_L^\top \left.{\partial d(z,y_s) \over \partial z}\right|_{z = \overline W_L \cdots \overline W_1 x_s} x_s^\top \overline W_1^\top \cdots \overline W_{k-1}^\top + 2\mu_k \overline W_k \\
& = 
 \overline W_{k+1}^\top \cdots \overline W_L^\top \left( \sum\nolimits_{s \in \mathcal I \cup \mathcal J} \left.{\partial d(z,y_s) \over \partial z}\right|_{z = \overline W_L \cdots \overline W_1 x_s} x_s^\top \right) \overline W_1^\top \cdots \overline W_{k-1}^\top + 2\mu_k \overline W_k.
\end{align*}
Since the first term has rank 1, so does $\overline W_k$, for each $k \in [L]$. \hfill $\blacksquare$

\section{Proof of Theorem \ref{app:prop-2}}
\label{appendix:prop-2}
For brevity in notation, define
\begin{align*}
\tilde x_{i} & = x_{i} \quad \ \ \forall i \in \mathcal I, \\
\tilde x_{j} & = -x_{j} \quad \forall j \in \mathcal J,
\end{align*}
and $\mathcal S = \mathcal I \cup \mathcal J$. Then the cost function (\ref{eqn:ce-loss}) could be written as
\[ \ell(W_1, \dots, W_L) = \sum\nolimits_{s \in \mathcal S} \log\left(1 + e^{W_L \cdots W_1 \tilde x_s}\right). \]

Note that at least one of the weight matrices must diverge to $\infty$ in norm since the loss function does not attain its minimum at a finite point. In addition, we have
\begin{align*} {d \|W_k(t) \|_F^2 \over dt }  & = 2 \left\langle W_k(t) , {d W_k(t) \over dt} \right\rangle \\
& = 2 \left\langle W_k(t) , - {\partial \ell \over \partial W_k} \right\rangle \\
& = {d \|W_{k'}(t)\|_F^2 \over dt} \quad \forall k, k' \in [L],
\end{align*}
and consequently,
\[ \|W_k(t)\|_F^2 - \|W_{k'}(t)\|_F^2 = \|W_k(0)\|_F^2 - \|W_{k'}(0)\|_F^2 \quad \forall t \ge 0,\ \forall k, k' \in [L].\]
Therefore, all of the weight matrices must diverge to $\infty$ in norm.

We can define the set of indices for the support vectors:
\begin{align*}
\mathcal S^\text{sup} & = \left\{ s \in \mathcal S : \lim_{t \to \infty} { \exp(W_L \cdots W_1 \tilde x_{s'})  \over \exp(W_L \cdots W_1 \tilde x_s) } < \infty  \ \forall s' \in \mathcal S \right\},
 \end{align*}
which is a nonempty set.
For any $k \in [L]$, remember that $\overline W_k \in \mathbb R^{n_k \times n_{k-1}}$. Let $i_1, i_2 \in [n_{k}]$ and $j_1, j_2 \in [n_{k-1}]$ be such that $e_{i_1}^\top \overline W_k e_{j_1}$, $e_{i_2}^\top \overline W_k e_{j_1}$ and $e_{i_2} \overline W_k e_{j_2}$ are nonzero. Note that if such a  tuple of $(i_1,i_2, j_1)$ does not exist, then the matrix $\overline W_k$ has rank 1 already, and the following analysis is not needed.
Given such a tuple of $(i_1, i_2, j_1)$, we have

\begin{align*}
{ e_{i_1}^\top \overline W_k e_{j_1} \over e_{i_2}^\top \overline W_k e_{j_1} } &  = \lim_{t \to \infty} { e_{i_1}^\top W_k(t) e_{j_1} \over e_{i_2}^\top W_k(t) e_{j_1} } \\
& =  \lim_{t \to \infty} { e_{i_1}^\top {d W_k \over dt} e_{j_1} \over e_{i_2}^\top {d W_k \over dt} e_{j_1} } \\
& = \lim_{t \to \infty} \frac{ \sum\nolimits_{s \in \mathcal S} e_{i_1}^\top  \left(  W_{k+1}^\top \cdots W_L^\top \tilde x_{s}^\top W_1^\top \cdots W_{k-1}^\top \right) e_{j_1} e^{ W_L \cdots W_1 \tilde x_{s}}}{
\sum\nolimits_{s \in \mathcal S} e_{i_2}^\top  \left(  W_{k+1}^\top \cdots W_L^\top \tilde x_{s}^\top W_1^\top \cdots W_{k-1}^\top \right) e_{j_1} e^{ W_L \cdots W_1 \tilde x_{s}}} \\
& = { e_{i_1}^\top \overline W_{k+1}^\top \cdots \overline W_L^\top \over e_{i_2}^\top \overline W_{k+1}^\top \cdots \overline W_L^\top } 
\lim_{t \to \infty} \frac{
\sum\nolimits_{s \in \mathcal S^\text{sup}}  \left(\tilde  x_s^\top \overline W_1^\top \cdots \overline W_{k-1}^\top \right) e_{j_1} e^{W_L \cdots W_1\tilde x_s - W_L \cdots W_1 \tilde x_{s_\text{sup}}} }{
\sum\nolimits_{s \in \mathcal S^\text{sup}}  \left(\tilde  x_s^\top \overline W_1^\top \cdots \overline W_{k-1}^\top \right) e_{j_1} e^{W_L \cdots W_1\tilde x_s - W_L \cdots W_1 \tilde x_{s_\text{sup}}} } \\
& = { e_{i_1}^\top \overline W_{k+1}^\top \cdots \overline W_L^\top \over e_{i_2}^\top \overline W_{k+1}^\top \cdots \overline W_L^\top } 
\lim_{t \to \infty} \frac{
\sum\nolimits_{s \in \mathcal S^\text{sup}}  \left(\tilde  x_s^\top \overline W_1^\top \cdots \overline W_{k-1}^\top \right) e_{j_2} e^{W_L \cdots W_1\tilde x_s - W_L \cdots W_1 \tilde x_{s_\text{sup}}} }{
\sum\nolimits_{s \in \mathcal S^\text{sup}}  \left(\tilde  x_s^\top \overline W_1^\top \cdots \overline W_{k-1}^\top \right) e_{j_2} e^{W_L \cdots W_1\tilde x_s - W_L \cdots W_1 \tilde x_{s_\text{sup}}} } \\
& =  \lim_{t \to \infty} { e_{i_1}^\top {d W_k \over dt} e_{j_2} \over e_{i_2}^\top {d W_k \over dt} e_{j_2} } \\
& = { e_{i_1}^\top \overline W_k e_{j_2} \over e_{i_2}^\top \overline W_k e_{j_2} } 
\end{align*}
where $\tilde x_{s_\text{sup}}$ denotes any point in the set $\{x_s\}_{s \in \mathcal S^\text{sup}}$. Note that this equality shows that all columns of $\overline W_k$ are in the span of the same single vector in $\mathbb R^{n_k}$, which proves that $\overline W_k$  has rank 1. \hfill $\blacksquare$

\section{Proof of Theorem \ref{theo:theo-4}}
\label{app:theo-4}

By duality of the norms $\|\cdot \|_p$ and $\|\cdot \|_q$, we have
\[ \min_{d : \|d\|_q \le \epsilon} w^\top (x_i + d) = w^\top x_i - \epsilon \|w\|_p,\]
\[ \max_{d : \|d\|_q \le \epsilon} w^\top (x_i + d) = w^\top x_i + \epsilon \|w\|_p. \]
Then problem (\ref{prob-2}) can be written as
\[ \min_w \sum_{i \in \mathcal I} {1\over 2} \left( y_i - w^\top x_i + \epsilon \|w\|_p \right)^2 + {1\over 2} \left( y_i - w^\top x_i - \epsilon \|w\|_p \right)^2, \]
which can be simplified to
\[ \min_{w} \sum_{i \in \mathcal I} \left( y_i - w^\top x_i \right)^2 + \epsilon^2 \|w\|_p^2 . \]
This is a convex problem in $w$, and we can introduce a slack variable to bring the second term into a constraint form:
\begin{align}
   \underset{w,t}{\text{minimize}} & \quad \sum\nolimits_{i \in \mathcal I} \left( y_i - w^\top x_i \right)^2 + \epsilon^2 t \label{aux-prob-1}\\
   \text{subject to} & \quad \|w\|_p^2 \le t. \nonumber
\end{align}
Fix $\epsilon > 0$, and assume that $(w_0, t_0)$ is the solution of (\ref{aux-prob-1}). 
Then $w_0$ is also a solution to the problem
\begin{align*}
    \underset{w}{\text{minimize}} & \quad \sum\nolimits_{i \in \mathcal I} (y_i - w^\top x_i)^2 \\
    \text{subject to} & \quad \|w\|_p^2 \le t_0,
\end{align*}
as well as
\begin{align}
    \underset{w}{\text{minimize}} & \quad \sum\nolimits_{i \in \mathcal I} (y_i - w^\top x_i)^2 \label{prob:aux} \\
    \text{subject to} & \quad \|w\|_p^m \le t_0^{m/2}. \nonumber
\end{align}
If $t_0 =0$, then $w_0$ is also zero, and this solution can be obtained by (\ref{prob-1}) by choosing $\lambda$ large enough. Therefore, without loss of generality assume $t_0 \neq 0$. Then problem (\ref{prob:aux}) satisfies the Slater's condition, and strong duality holds \citep{boyd2004convex}. Then we can find its solution by solving 
\[ \min_{w} \sum\nolimits_{i \in \mathcal I} (y_i - w^\top x_i)^2 + \lambda^*( \|w\|_p^m - t_0^{m/2}) \]
where $\lambda^*$ is the dual solution. Note that this problem is strictly convex in $w$, and therefore, its solution is unique, for which the only candidate is $w_0$. We conclude that
\[ w_0 = \argmin_w \sum\nolimits_{i\in \mathcal I} (y_i - w^\top x_i)^2 + \lambda^* \|w\|_p^m. \]
This completes the one direction of the proof, and the other direction is identical. \hfill $\blacksquare$

\end{document}